%% file: arxiv.tex
\documentclass[runningheads]{llncs}
\usepackage{graphicx}
\usepackage{amsmath,amssymb} 
\usepackage{color}
\usepackage{xcolor}
\usepackage[width=122mm,left=12mm,paperwidth=146mm,height=193mm,top=12mm,paperheight=217mm]{geometry}
\usepackage{ragged2e}
\usepackage{epsfig}
\usepackage{array}
\authorrunning{Scene Text Image Super-Resolution in the Wild}
\usepackage{algorithm}
\usepackage{color}

\usepackage{algpseudocode}
\usepackage{subfigure}

\usepackage{multirow}
\usepackage[breaklinks=true,bookmarks=false]{hyperref}

\begin{document}
\title{Scene Text Image Super-Resolution in the Wild} 

\author{Wenjia Wang\inst{1*} \and
Enze Xie\inst{2}\thanks{Equal Contribution.} \and
Xuebo Liu\inst{1} \and \\
Wenhai Wang\inst{3} \and 
Ding Liang\inst{1} \and 
Chunhua Shen\inst{4} \and
Xiang Bai\inst{5}}

\institute{SenseTime Research \and The University of Hong Kong \and Nanjing University \and The University of Adelaide \and Huazhong University of Science and Technology \\}
\maketitle



\begin{abstract}

Low-resolution text images are often seen in natural scenes such as documents captured by mobile phones.
Recognizing low-resolution text images is challenging because they lose detailed content information, leading to poor recognition accuracy. 
An intuitive solution is to introduce super-resolution~(SR) techniques as pre-processing. However, previous single image super-resolution~(SISR) methods are trained on synthetic low-resolution images (\emph{e.g.} Bicubic down-sampling), which is simple and not suitable for real low-resolution text recognition. 
To this end, we propose a real scene text SR dataset, termed TextZoom. It contains paired real low-resolution and high-resolution images which are captured by cameras with different focal length in the wild. 
It is more authentic and challenging than synthetic data.
A new Text Super-Resolution Network, termed TSRN, with three novel modules is developed.
(1)~A sequential residual block is proposed to extract the sequential information of the text images. 
(2)~A boundary-aware loss is designed to sharpen the character boundaries.
(3)~A central alignment module is proposed to relieve the misalignment problem in TextZoom.
Extensive experiments on TextZoom demonstrate that our TSRN largely improves the recognition accuracy by over 13\% of CRNN, and by nearly 9.0\% of ASTER and MORAN compared to synthetic SR data.  
Furthermore, our TSRN clearly outperforms 7 state-of-the-art SR methods in boosting the recognition accuracy of LR images in TextZoom. 
Our results suggest that low-resolution text recognition in the wild is far from being solved, thus more research effort is needed.
The codes and models will be released at: \href{https://github.com/JasonBoy1/TextZoom}{\color{blue} \tt github.com/JasonBoy1/TextZoom} 

\keywords{Scene Text Recognition, Super-Resolution, Dataset, Sequence, Boundary}
\end{abstract}

\begin{figure}[ht]
    \centering
    \scalebox{1}{\includegraphics[width=1\textwidth]{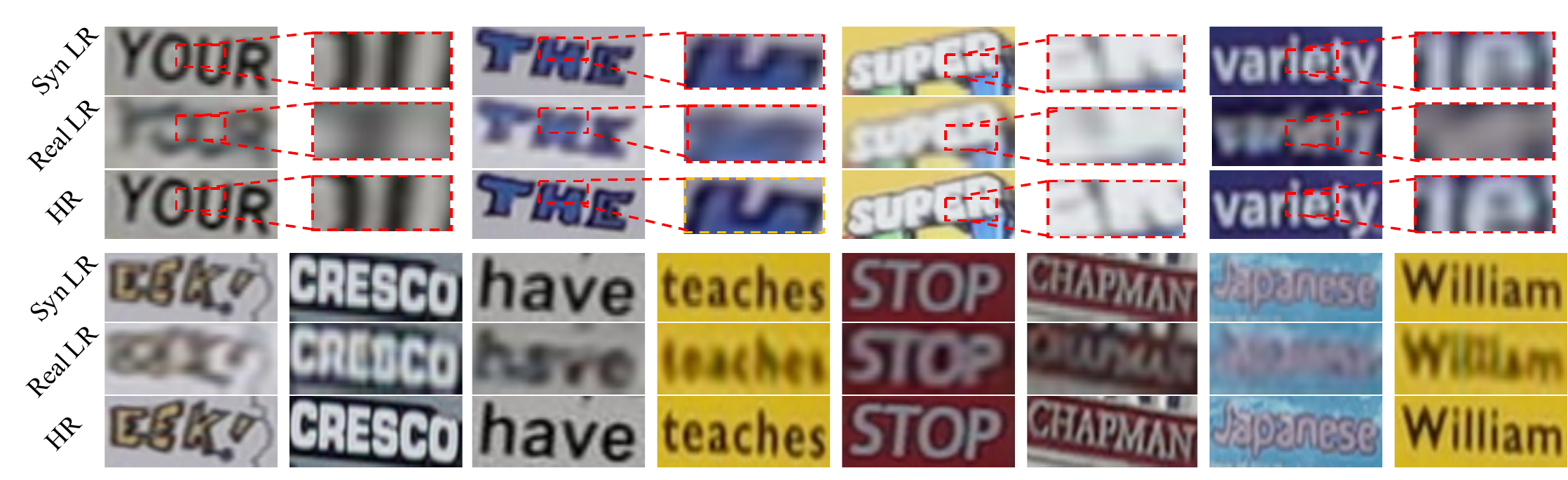}}
    
    \caption{Comparison between synthetic LR, real LR, and HR images in TextZoom. `Syn LR' denotes BICUBIC down-sampled image of HR. `Real LR' and `HR' denotes LR and HR images captured by camera with different focal lengths. From the images we can find that the real LR images are much more challenging than the synthetic LR images.
    }
    \label{fig:real_syn}
\end{figure}

\begin{table}[ht]
\centering

	\caption{Statistics of TextZoom. The testing set is divided into 3 different subsets: easy, medium and hard. The recognition accuracy is tested by ASTER~\cite{aster}. We see the recognition accuracy of LR images decreases when the difficulty increases. Our main purpose is to increase the recognition accuracy of the LR images by super-resolution.}
	\label{tab:textzoom}

    \scalebox{1}{
    \input{tables/textzoom_allocation.tex}
    }

\end{table}

\begin{figure}[ht]
    \centering
    \scalebox{0.85}{\includegraphics[width=1\textwidth]{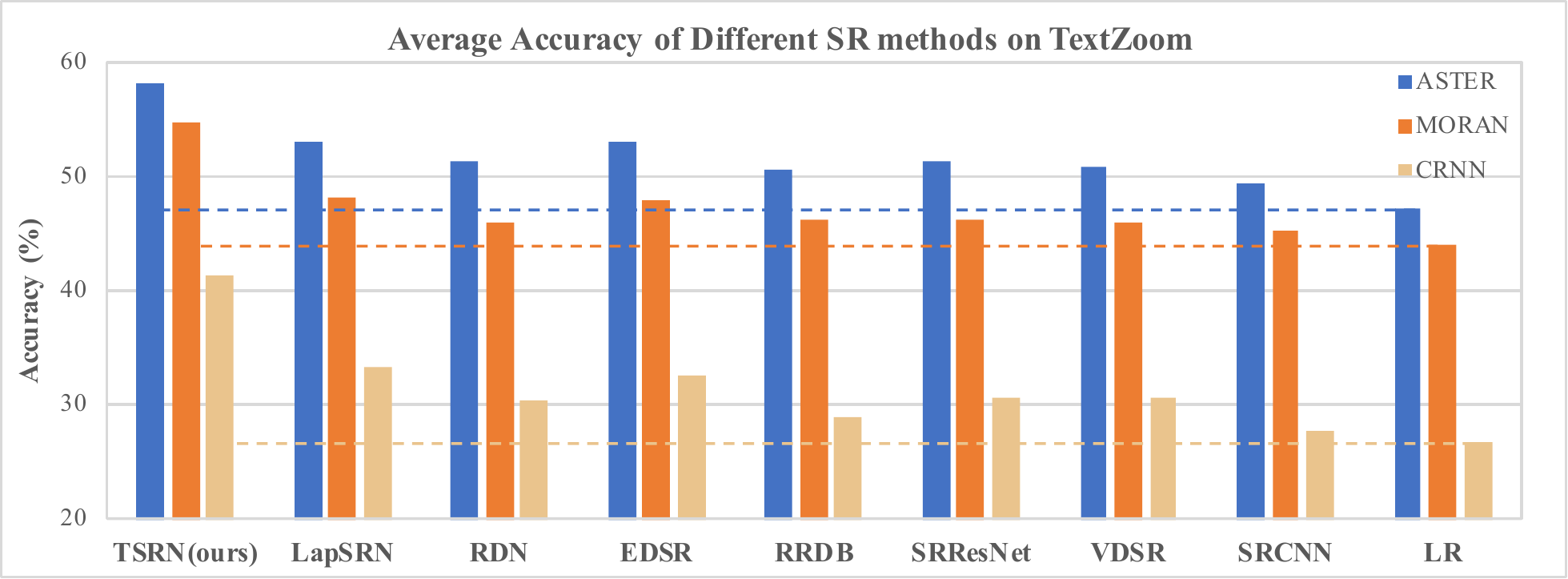}}

    \caption{Average recognition accuracy of the super-resolved images of LR images in TextZoom. We first super-resolve LR images with different SR methods, then directly test the SR results with the official released model of ASRER~\cite{aster}, MORAN~\cite{luo2019moran} and CRNN~\cite{crnn}. We compare our TSRN with 7 state-of-the-art deep learning networks and show ours outperforms them clearly. Dotted lines means accuracy of LR inputs.
    }
    \label{fig:aster_moran_crnn}
 
\end{figure}

\section{Introduction}
Scene text recognition is a fundamental and important task in computer vision, since it is usually a key step towards many downstream text-related applications, including document retrieval, card recognition, license plate recognition, etc~\cite{handwriPR,ic19document,chinesehandw,licenseplate}. Scene Text recognition has achieved remarkable success due to the development of Convolutional Neural Network~(CNN).

Many accurate and efficient methods have been proposed for most constrained scenarios (e.g., text in scanned copies or network images).
Recent works focus on texts in natural scenes~\cite{liu2016star,liu2018squeezedtext,cheng2017focusing,luo2019moran,aster,spcnet,psenet,pan}, 
which is much more challenging due to the high diversity of texts in blur, orientation, shape, and low-resolution. A thorough survey of recent advantages of text recognition can be found in~\cite{long2018scene} .
Modern text recognizers have achieved impressive results on clear text images.
However, their performances drop sharply when recognizing low-resolution text images~\cite{whatswrong}. 
The main difficulty to recognize LR text is that the optical degradation blurred the shape of the characters. \textbf{Therefore, it would be promising if we introduce SR methods as a pre-processing procedure before recognition.} To our surprise, none of the real dataset and corresponding methods focus on scene text SR.

In this paper, we propose a paired scene text SR dataset, termed TextZoom, which is the \textbf{first dataset focus on real text SR}. Previous Super-Resolution methods~\cite{srcnn,vdsr,srgan,edsr,esrgan,rdn,lapsrn}
 generate LR counterparts of the high-resolution (HR) images by simply applying uniform degradation like bicubic interpolation or blur kernels. Unfortunately, real blur scene text images are more varied in degradation formation.
Scene texts are of arbitrary shapes, distributed illumination, and different backgrounds. Super-resolution on scene text images is much more challenging. Therefore, the proposed TextZoom, which contains paired LR and HR text images of the same text content, is very necessary. The TextZoom dataset is cropped from the newly proposed SISR datasets ~\cite{realsr,srraw}. Our dataset has three main advantages. \textbf{(1)} This dataset is well annotated. We provide the direction, the text content and the original focal length of the text images. \textbf{(2)} The dataset contains abundant text from different natural scenes, including street views, libraries, shops, vehicle interiors and so on. \textbf{(3)} The dataset is carefully divided into three subsets by difficulty. 
Experiments on TextZoom demonstrate that  our  TSRN  largely  improves  the  recognition  accuracy  of  CRNN by over 13\% compared to synthetic SR data.
The annotation and allocation strategy will be briefly introduced in section~\ref{sec:dataset} and demonstrated in detail in supplementary materials.

 Moreover, to reconstruct low-resolution text images, we propose a text-oriented end-to-end method. 
 Traditional SISR methods only focus on reconstruct the detail of texture and only satisfy human's visual perception.
 However, scene text SR is quite a special task since it contains high-level text content. The fore-and-aft characters have information relations with each other. Obviously, a single blur character will not disable human to recognize the whole word if other characters are clear. To solve this task, firstly, we present a Sequential Residual Block to model recurrent information in text lines, which enabling us to build a correlation in the fore-and-aft characters. Secondly, we propose a boundary-aware loss termed gradient profile loss to reconstructing the sharp boundary of the characters. This loss helps us to distinguish between the characters and backgrounds better and generate a more explicit shape. Thirdly, the misalignment of the paired images is inevitable due to the inaccuracy of the cameras.  We propose a central alignment module to make the corresponding pixels more aligned. 
 We evaluate the recognition accuracy by two steps:
\textbf{(1)} Do super-resolution with different methods on LR text images;
\textbf{(2)} Evaluate the SR text images with trained Text Recognizers~\emph{e.g.} ASTER, MOCAN and CRNN. 
 Extensive experiments show our TSRN clearly outperforms 7 state-of-the-art SR methods in boosting the recognition accuracy of LR images in TextZoom. For example, it outperforms LapSRN by over  5\%  and  8\%  on  recognition  accuracy  of  ASTER  and CRNN. Our results suggest that low-resolution text recognition in the wild is far from being solved, thus more research effort is needed.

The contributions of this work are therefore three-fold: 
\begin{enumerate}
    \item We introduce the first \textbf{real} paired scene text SR dataset TextZoom with different focal lengths. We annotate and allocate the dataset with three subsets: easy, medium and hard, respectively. 

    \item We prove the superiority of the proposed dataset TextZoom by comparing and analyzing the models trained on synthetic LR and proposed LR images. We also prove the necessity of scene text SR from different aspects.
    \item We propose a new text super-resolution network with three novel modules. It surpasses 7 representative SR methods clearly by training and testing them on TextZoom for fair comparisons.
\end{enumerate}

\section{Related work}

\textbf{Super-Resolution.} 
Super-resolution aims to output a plausible high-resolution image that is consistent with a given low-resolution image. Traditional approaches, such as bilinear, bicubic or designed filtering, leverage the insight that neighboring pixels usually exhibit  similar colors and generate the output by interpolating between the colors of neighboring pixels according to a predefined formula. 
In the deep learning era, super-resolution is treated as a regression problem, where the input is the low-resolution image, and the target output is the high-resolution image~\cite{srcnn,vdsr,srgan,esrgan,edsr,rdn,lapsrn}. A deep neural net is trained on the input and target output pairs to minimize some distance metric between the prediction and the ground truth. 
These works are mainly trained and evaluated on those popular datasets~\cite{2012set5,2010set15,BSD300,urban100,manga109,div2k}. In these datasets, LR images are generated by a down-sample interpolation or Gaussian blur filter. 
Recently, several works capture LR-HR images pairs by adjusting the focal length of the cameras~\cite{realsr,srraw,camerasr}.
In~\cite{realsr,camerasr}, a pre-processing method is applied to reduce the misalignment between the captured LR and HR images While in~\cite{srraw}, a contextual bilateral loss is proposed to leverage the misalignment. In this work, a new dataset TextZoom is proposed, which fills in the absence of paired scene text SR dataset. It is well annotated and allocated with difficulty. We hope it can serve as a challenging benchmark.

\textbf{Text Recognition.}
Early work adopts a bottom-up fashion~\cite{jaderberg2014} which detects individual characters firstly and integrates them into a word, or a top-down manner~\cite{readingijcv}, which treats the word image patch as a whole and recognizes it as a multi-class image classification problem.  
Considering  that  scene text generally appears as a character sequence, CRNN~\cite{crnn} regard it as a sequence recognition problem and employs Recurrent Neural Network (RNNs) to model the sequential features. CTC~\cite{ctc} loss is often combined with the RNN outputs for calculating the conditional probability between the predicted sequences and the target~\cite{liu2016star,liu2018squeezedtext}.  
Recently, an increasing number of recognition approaches based on the attention mechanism have achieved significant improvements~\cite{cheng2017focusing,luo2019moran}. 
ASTER~\cite{aster} rectified oriented or curved text based on Spatial Transformer Network(STN)~\cite{stn} and then performed recognition using an attentional sequence-to-sequence  model. In this work, we choose state-of-the-art recognizer ASTER~\cite{aster}, MORAN~\cite{luo2019moran} and CRNN~\cite{crnn} as baseline recognizers to evaluate the recognition accuracy of the SR images.

\textbf{Scene Text Image Super-Resolution.}
Some previous works conducted on scene text image super-resolution are aimed at improving the recognition accuracy and image quality evaluation metrics. ~\cite{introductiontsr} compared the performance of several artificial filters on down-sampled text images. ~\cite{bidocumentsr} propose a convolution-transposed convolution architecture to deal with binary document SR. ~\cite{dong2015boosting} adapt SRCNN~\cite{srcnn} in text image SR in the ICDAR 2015 competition TextSR~\cite{textsr2015}  and achieved a good performance, but no text-oriented method was proposed. 

These works take a step on low-resolution text recognition, but they only train on down-sampled images, learning to regress a simple mapping function of inverse-bicubic (or bilinear) interpolation. Since all the LR images are identically generated by a simple down-sample formulation, it is not well-generalized to real text images.

\section{TextZoom Dataset}
\label{sec:dataset}

\textbf{Data Collection \& Annotation.}
Our proposed dataset TextZoom comes from two state-of-the-art SISR datasets: RealSR~\cite{realsr} and SRRAW~\cite{srraw}.
These two newly proposed datasets consist of paired LR-HR images captured by digital cameras.

RealSR~\cite{realsr} is captured by four focal lengths with two digital cameras: Canon 5D3 and Nikon D810. In RealSR~\cite{realsr}, these four focal lengths of images are allocated as ground truth, 2X LR images, 3X LR images, 4X LR images separately. For RealSR, we annotate the bounding box of the words on the 105mm focal length images. SR-RAW is collected by seven different focal lengths with SONY FE camera, range from 24-240mm. 
The images captured in shorted focal lengths could be used as LR images while those captured in longer lengths as corresponding ground truth. 
For SR-RAW, we annotate the bounding box of the words on the 240mm focal length images.

We labeled the images with the largest focal length of each group and cropped the text boxes from the rest following the same rectangle. So the misalignment is unavoidable. There are some top-down or vertical text boxes in the annotated results. In this task, we rotate all of these images to horizontal for better recognition. There are only a few curved text images in our dataset.
For each pair of LR-HR images, we provide the annotation of the case sensitive character string (including punctuation), the type of the bounding box, and the original focal lengths.
We demonstrate the detailed annotation principle of the text images cropped from SR-RAW and RealSR in detail in supplementary materials.

The size of the cropped text boxes is diverse, \emph{e.g.} height from 7 to 1700 pixels, so it is not suitable to treat the text images cropped from the same focal lengths as a same domain. We define our principle following these considerations.
 \textbf{(1)No patching.} In SISR, data are usually generated by cropping patches from the original images~\cite{srgan,esrgan,rcan,realsr,srraw}.
Text images could not be cut into patches since the shape of the characters should maintain completed. 
\textbf{(2) Accuracy distribution.} We divide the text images by height and test the accuracy (Refer to the Tables showed in supplementary materials). We found that the accuracy does not increase obviously when the height is larger than 32 pixels. 
Setting images to 32 pixels height is also a customary rule in scene text recognition research~\cite{crnn,cheng2017focusing,luo2019moran}. The accuracy of the images smaller than 8 pixels are too low, which hardly has any value for super-resolution, so we discard the images the height of which is less than 8 pixels. 
\textbf{(3) Number.} We found that in the cropped text images, the height range from 8 to 32 claim the majority. \textbf{(4) No down-sample.} Since the interpolation degradation should not be introduced into real blur images, we could only up-sample the LR images to a relatively bigger size.

Following these 4 considerations, we up-sample the images ranging from 16-32 pixels height to 32 pixels height, and up-sample the images ranging from 8-16 pixels height to 16 pixels height. We conclude that (16, 32) should be a good pair to form a 2X train set for scene text SR task. For example, the text images taken from 150mm focal length and height sized in 16-32 pixels would be taken as a ground truth for the 70mm counterpart. So we selected all the images the height of which range from 16 pixels to 32 pixels as our ground truth image and up-sample them to the size of 128$\times$32~(width$\times$height), and the corresponding 2X LR images to the size of 64$\times$16~(width$\times$height).  
For this task, we only generate this 2X LR-HR pair dataset from the annotated text images mainly due to the special characteristics of text recognition. Other scale of factors of our annotated images could be used for different purpose.

\textbf{Allocation of TextZoom.}
The SR-RAW and RealSR are collected by different cameras with different focal lengths. The distance from the objects also affect the legibility of the images. So the dataset should be further divided following their distribution.
 
The train-set and test-set are cropped from the original train-set and test-set in SR-RAW and RealSR separately. 
The author of SR-RAW used larger distance from the camera to the subjects to minimize the perspective shift~\cite{srraw}. So the accuracy of text images from SR-RAW is relatively lower under the similar focal lengths compared to RealSR. The accuracy of the images cropped from 100mm focal lengths in SR-RAW is 52.1\% tested by ASTER~\cite{aster}, while the accuracy of those from 105mm in RealSR is 75.0\% tested by ASTER~\cite{aster} (Refer to the Tables showed in supplementary materials). With the same height, the images of smaller focal lengths are more blurred. With this in mind, we allocate our dataset into three subsets by difficulty. The LR images cropped from RealSR render \textbf{easy}. The LR images from SR-RAW and the focal lengths of which larger than 50mm are viewed as \textbf{medium}. The rest are as \textbf{hard}.

In this task, our main purpose is to increase the \textbf{recognition accuracy} of the easy, medium and hard subsets. We also show the results of peak signal to noise ratio (PSNR) and structural similarity index (SSIM) in the supplementary materials.

\textbf{Dataset Statistics}
The detailed statistics of TextZoom is shown in supplementary materials.

\section{Method}
In this section, we present our proposed method TSRN in detail.
Firstly, we briefly describe our pipeline in section~\ref{pipeline}. Then we demonstrate the proposed Sequential Residual Block. Thirdly, we introduce our central alignment module. Finally, we introduce a new gradient profile loss to sharpen the text boundaries.

\subsection{Pipeline}
\label{pipeline}

\begin{figure}[ht]
\centering
\includegraphics[width=0.91\textwidth]{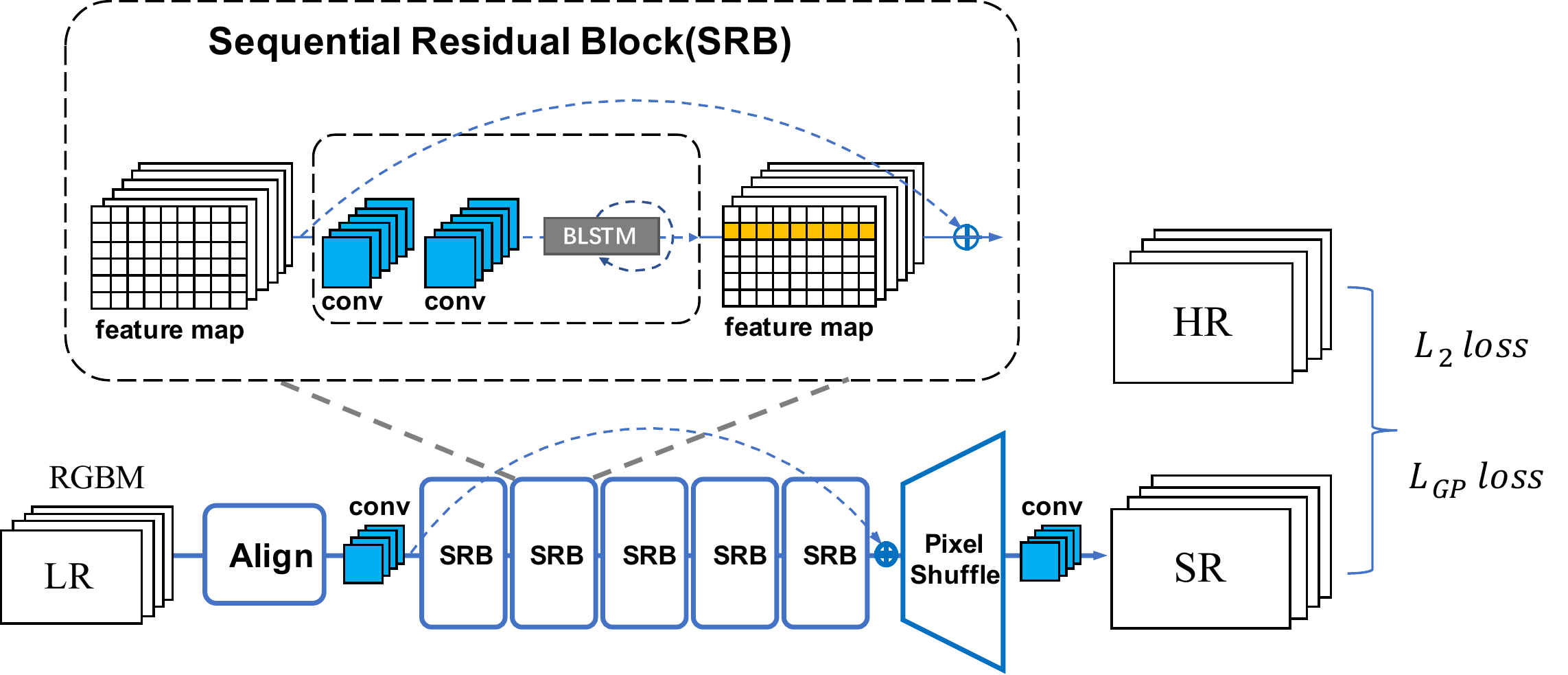}

\caption{The illustration of our proposed TSRN. We concatenate binary mask with RGB channels as a RGBM 4-channel input. The input is recitified by central alignment module and then fed into our pipeline. The output is the super-resolved RGB image. The outputs are supervised by $L_{2}~ loss$. The RGB channels of the outputs are supervised by $L_{GP}~loss$.}
\label{fig:pipeline}

\end{figure}

Our baseline is SRResNet~\cite{srgan}. As shown in Fig.~\ref{fig:pipeline}, we mainly make two modifications to the structure of SRResNet: 1) add a central alignment module in front of the network; 2) replace the original basic blocks with the proposed Sequential Residual Blocks (SRBs). In this work, we concatenate the binary mask with RGB image as our input. The binary masks are simply generated by calculating the mean gray scale of the image. The detailed information of masks is shown in supplementary materials. During training, firstly, the input is rectified by central alignment module. Then we use CNN layers to extract shallow features from the rectified image. Stacking five SRBs, we extract deeper and sequential dependent feature and do shortcut connection following ResNet~\cite{resnet}. The SR images are finally generated by up-sampling block and CNN. 
We also design a gradient prior loss ($L_{GP}$) aiming at enhancing the shape boundary of the characters. The output of the network is supervised by MSELoss ($L_{2}$) and our proposed gradient profile loss ($L_{GP}$).

\subsection{Sequential Residual Block}
Previous state-of-the-art SR methods mainly pursue better performance in PSNR and SSIM.
Traditional SISR only cares about texture reconstruction while text images have strong sequential characteristics. 
In text recognition tasks, scene text images encode the context information for text recognition by Recurrent Neural Network (RNN)~\cite{readscenetext}. 
Inspired from them, we modified the residual blocks~\cite{srgan} by adding Bi-directional LSTM (BLSTM) mechanism. Inspired by~\cite{ctpn}, we build sequence connectionist in horizontal lines and fused the feature into deeper channels. Different from~\cite{ctpn}, we build the in-network recurrence architecture not for detecting but for low-level reconstruction, so we only adapt the idea of building text line sequence dependence. In Fig.~\ref{fig:pipeline}, the SRB is briefly illustrated. Firstly, we extract feature by CNN. Then permute and resize the feature map as the horizontal text line can be encoded into sequence. Then the BLSTM can propagate error differentials~\cite{crnn}, and invert the feature maps into feature sequences, and feed them back to the convolutional layers.
To make the sequence dependent robust for tilted text images, we introduce the BLSTM from two directions, horizontal and vertical. 
BLSTM takes the horizontal and vertical convolutional feature as sequential inputs, and updates its internal state recurrently in the hidden layer.

\begin{equation}
\begin{aligned}
    H_{t_{1}}=\phi_{1}(X_{t_{1}}, H_{t_{1}-1}),
    ~~~~t_{1}=1,2,...,W\\
        H_{t_{2}}=\phi_{1}(X_{t_{2}}, H_{t_{2}-1}),
    ~~~~t_{2}=1,2,...,H
\end{aligned}
\end{equation}
Here $H_{t}$ denotes the hidden layers, $X_{t}$ denotes the input features, $t_{1}, t_{2}$ separately denote the recurrent connection from horizontal and vertical direction.

\subsection{Central Alignment Module}
The misalignment make the pixel-to-pixel losses, such as $L_{1}$ and $L_{2}$ generate significant artifacts and double shadows. This mainly due to the misalignment of the pixels in training data. Sine some of the text pixels in LR images are in spatial corresponding to the background pixels in the HR images, the network could learn a wrong pixel-wise counterpart information.
As mentioned in Section.~\ref{sec:dataset}, the text regions in HR images are more central aligned compared to the LR images. So we introduce STN\cite{stn} as our central alignment module. The STN is a spatial transform network which can rectify the images and
be learned end-to-end. To rectify spatial variation flexibly, we adopt TPS transformation as the transform manipulation. Once the text regions in LR images are aligned adjacent the center, the pixel-wise losses would make better performance and the artifacts could be relieved. We show more detailed information of central alignment module in supplementary materials.

\subsection{Gradient Profile Loss}
Gradient Profile Prior~(GPP) is proposed in~\cite{GPP} to generate sharper edge in SISR task. 
Gradient field means the spatial gradient of the RGB values of the pixels.

Since we have a paired text super-resolution dataset, we could use the gradient field of HR images as ground truth.
Generally, the color of characters in text images contrast strongly with the backgrounds.So sharpening the boundaries rather than smooth ones of characters could make the characters more explict.

\begin{figure}[ht]
\centering

\includegraphics[width=1\textwidth]{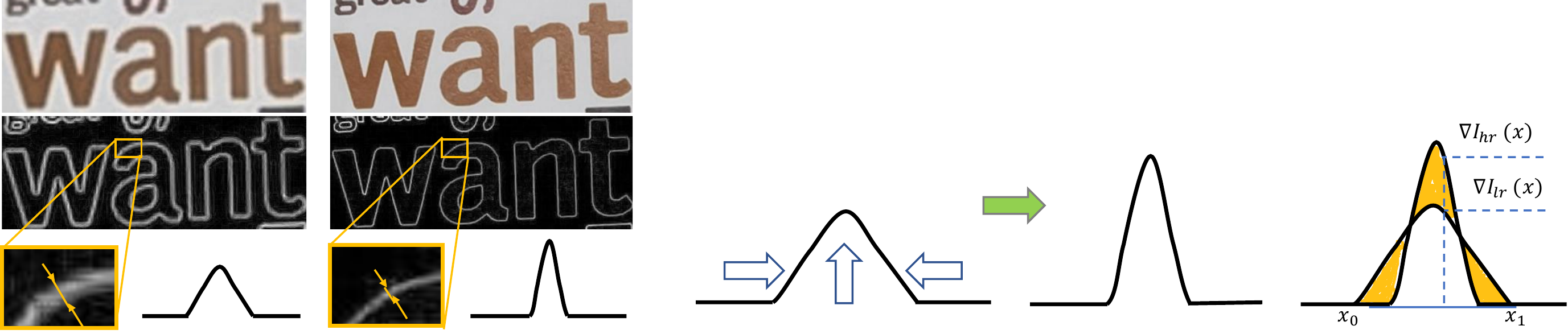}

\caption{The illustration of gradient field and Gradient Prior Loss.}
\label{fig:gradient}
\end{figure}

We revisit the GPP and generate ground truth from HR images, then we define the loss function as below:
\begin{equation}
   \begin{aligned}
   L_{GP} = \mathbb{E}_{x}\vert \vert \nabla I_{hr}(x) - \nabla I_{sr}(x) \vert \vert_{1}   ~~~~~(x\in[x_{0}, ~x_{1}])
   \end{aligned}
   \label{lgp}
\end{equation}
$\nabla I_{hr}(x)$ denotes the gradient field of HR images, and $\nabla I_{sr}(x)$ denotes that of SR images. 

Our proposed $L_{GP}$ exhibits two advantageous properties: (1) The gradient field vividly show the characteristics of text images: the texts and backgrounds. (2) The LR images always come with wider curve of gradient field, while HR images mean thinner curve. And the curve of gradient field could be easily generated through
mathematical calculation. This ensures a confidential supervision label.

\section{Experiments}

\subsection{Datasets}
We train the SR methods on our proposed TextZoom (see section~\ref{sec:dataset}.) training set. We evaluate our models on our three subsets \textbf{easy}, \textbf{medium} and \textbf{hard}.
To avoid down-sample degradation, all the LR images are up-sampled to 64$\times$16, and HR images to 128$\times$32. 

\subsection{Implementation Details}
During training, we set the trade-off weight of $L_{2}$ loss as 1 and $L_{GP}$ as 1$e-4$. We use the Adam optimizer with momentum term 0.9. 
When evaluating recognition accuracy, we use the official Pytorch version code and the released model of ASTER: \href{https://github.com/ayumiymk/aster.pytorch}{\color{blue} \tt aster.pytorch}, MORAN: \href{https://github.com/Canjie-Luo/MORAN\_v2}{\color{blue} \tt MORAN\_v2.pytorch}, CRNN: \href{https://github.com/meijieru/crnn.pytorch}{\color{blue} \tt crnn.pytorch} from github.

All the SR models are trained by 500 epochs with 4 NVIDIA GTX 1080ti GPUs. The batch-size is adapted as the setting in the original papers.

\subsection{Is SR necessary for Text Recognition?}
\label{necessity}
We further quantitatively analyzed the necessity of super-resolution from three aspects.

It is assumed that we could achieve better performance on recognizing low-resolution (LR) text images if we directly train the recognition networks on small size images, and then the super-resolution procedure could be removed. This query is reasonable because the deep neural networks have a strong robustness on the training domains. To refute this query and prove the necessity of super-resolution for text images, we compare the recognition accuracy of 4 methods:
\begin{itemize}
    \item \textbf{Released.} Recognize with ASTER~\cite{aster} model trained on customary size (no less than 32 pixels in height, We use official released model here).
    \item  \textbf{ReIm.} Recognize with model trained on low-resolution images (In this work, we re-implemented ASTER~\cite{aster} on Syn90K~\cite{syn90k} and SynthText~\cite{syn800k} at the size of 64$\times$16, All the training details are the same as the original paper except the input sizes
  \item\textbf{Fine-tune.} Fine-tune released ASTER~\cite{aster} model on our TextZoom training set.
  \item\textbf{Ours.} Choose the low resolution images by size, then use our proposed TSRN to generate the SR images and then recognize them with ASTER~\cite{aster} official released model.
      ).
  
\end{itemize}

\begin{table}[ht]
\center 

\caption{Comparison between different methods. \textbf{Released} means official released model from github. \textbf{ReIm} means our re-implemented model trained on Syn90K~\cite{syn90k} and SynthText~\cite{syn800k} at the size of 64$\times$16.
}
	\label{tab:necessity}
\scalebox{0.9}{
    \input{tables/necessity.tex}
    }

\end{table}
To verify the robustness, we select all the images smaller than 64$\times$16 from 7 common scene text testing sets, IC13, IC15, CUTE, IC03, SVT, SVTP, CUTE and IIIT5K and get 436 images in total. We term this testing set \textbf{CommonLR}. 
We compare these 4 methods on our dataset TextZoom and CommonLR.
From Table~\ref{tab:necessity}, we can figure that the re-implemented model do increase the accuracy sharply on the LR images. The average accuracy of TextZoom can be increased by 5.4\%, from 47.2\% to 52.6\%. And the accuracy of CommonLR could also be improved for 5\%. 
The result of re-implemented model is still lower than the accuracy of our results (TSRN(ours) + ASTER(Released)). 

When we fine-tune the Aster on our TextZoom training set, the accuracy of TextZoom testing set would be even higher than our method. But TextZoom is a small sized dataset for recognition task, its different distribution would make the recognizer over-fit on it. The accuracy of CommonLR of fine-tune method is the lowest. Moreover, on this fine-tune Aster model the other testing sets like IC13, IC15, etc. would drop sharply for more than 10.0\% points.

Actually, our method is superior to fine-tune and re-Im methods in following aspects.
(1). The fine-tuned model over-fit on TextZoom. It achieves highest performance on TextZoom while lowest on CommomLR because the number of TextZoom is far from enough for text recognition task. 
Super-resolution ,a low-level task, usually needs less data to converge. Our method could directly choose SR or not by the size and get better overall performance.

(2).Our SR method can also produce better visual results for people to read (see Fig.~\ref{fig:display_ablation}). 
(3).While re-Im and fine-tune method need 2 recognition models for big and small size images separately, 
our method only need a tiny SR model, introducing marginal computation cost.  This part could be found in supplementary materials.

So the SR methods could be a effective and convenient pre-processing procedure of scene text recognition.

\subsection{Synthetic LR vs. TextZoom LR}
To demonstrate the superiority of paired scene text SR images, we compare the performance of the models trained on synthetic datasets and our TextZoom dataset. The quantitative results are shown in the supplementary materials.

\subsection{Ablation Study on TSRN}

\begin{table}[ht]
\center
\caption{Ablation study for different settings of our method TSRN. The recognition accuracies are tested by the official released model of ASTER~\cite{aster}.}

\scalebox{0.75}{
    \input{./tables/ablation_simple}
    }
	\label{tab:ablation}

\end{table}

In order to study the effect of each component in TSRN, we gradually modify the configuration of our network and compare their differences to build a best network. For brevity, we only compare the accuracy of ASTER~\cite{aster}.

\begin{figure}[ht]
    \centering
    \includegraphics[width=0.98\textwidth]{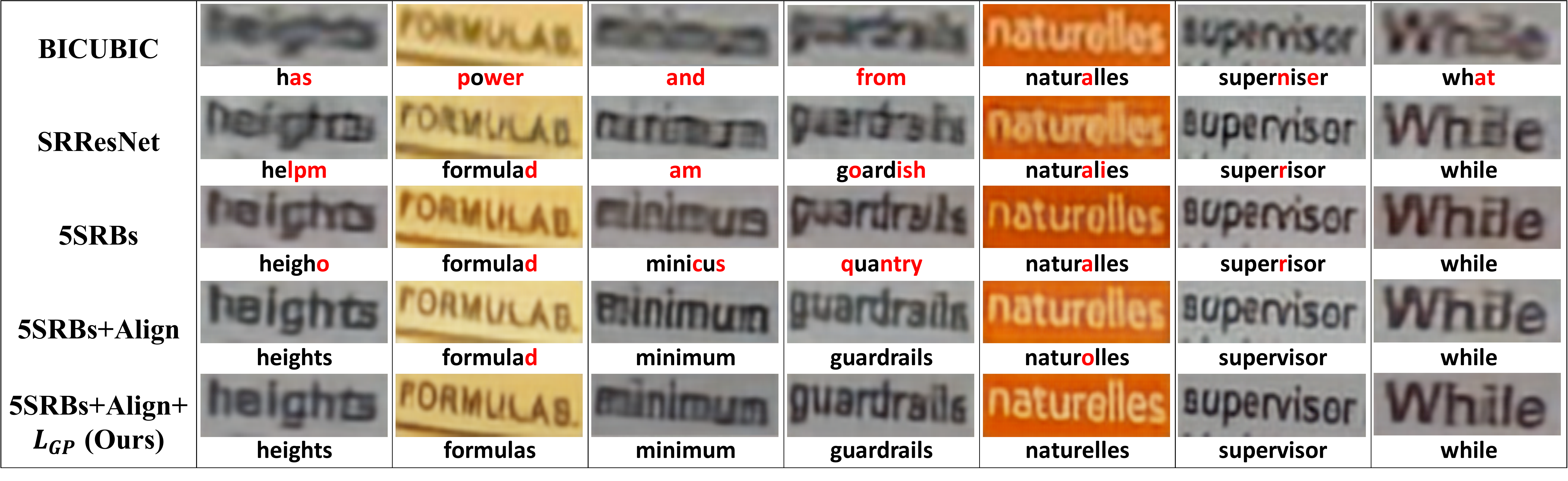}
    \caption{Visual comparisons for showing the effects of each component in our proposed TSRN. The recognition result strings of ASTER are displayed under each image. Those characters in red denote wrong recognition.}
    \label{fig:display_ablation}
\end{figure}

\textbf{1) SRBs.}
We add BLSTM mechanism to the basic residual block in SRResNet~\cite{srgan} and get the proposed SRB. The SRB is the essential component in TSRN. Comparing $\#$ 0 and $\#$ 1 in Table~\ref{tab:ablation} , stacking 5 SRBs, we can boost up the average accuracy by 4.9\% compared to SRResNet~\cite{srgan}.

\textbf{2) Central Alignment Module.}
Central alignment module can boost the average accuracy by 1.5\%, as shown in Table~\ref{tab:ablation} method 2. From Fig.~\ref{fig:display_ablation}, we can find that without central alignment module, the artifacts are strong, and the characters are twisted. While with more appropriate alignment, we could generate higher quality images since the pixel-wise loss function could supervise the training better.

\textbf{3) Gradient Profile Loss.}
From Table~\ref{tab:ablation} method 3, we can find the proposed gradient profile loss can boost the average accuracy by 0.5\%. Although the increase is slight, the visual results are better (Fig.~\ref{fig:display_ablation} method 3). 

In supplementary materials, we further discuss about the detailed component of our method.

\subsection{Comparison with State-of-the-Art SR methods}

\begin{table*}[ht]
\centering
\caption{Performance of state-of-the-art SR methods on the three subsets in TextZoom. For better displaying, we calculated the average accuracy. $L_{1}$ denotes Mean Average Error (MAE) Loss. $L_{2}$ denotes Mean Squared Error (MSE) Loss. $L_{tv}$ denotes Total Variation Loss. $L_{p}$ denotes Perceptual Loss proposed in \cite{johnson2016perceptual}. $Charbonnier$ denotes the Charbonnier Loss proposed in LapSRN~\cite{lapsrn}. $L_{GP}$ denotes our proposed Gradient Prior Loss.
The recognition accuracies are tested by the official released model of ASTER~\cite{aster}, MORAN~\cite{luo2019moran} and CRNN~\cite{crnn}.}

\scalebox{0.64}{
    \input{tables/main.tex}
    }
  	\label{tab:main}
\end{table*}

To prove the effectiveness of TSRN, we compare it with 7 SISR methods on our TextZoom dataset, including SRCNN~\cite{srcnn}, VDSR~\cite{vdsr}, SRResNet~\cite{srgan}, RRDB~\cite{esrgan}, EDSR~\cite{edsr}, RDN~\cite{rdn} and LapSRN~\cite{lapsrn}.
All of the networks are trained on our TextZoom training set and evaluated on our three testing subsets.

In Table~\ref{tab:main}, we list the recognition accuracy tested by ASTER~\cite{aster}, MORAN~\cite{luo2019moran}, and CRNN~\cite{crnn} of all the mentioned 7 methods, along with BICUBIC and the proposed TSRN. In Table~\ref{tab:main}, it can be observed that TSRN outperforms all the 7 SISR methods in recognition accuracy sharply. Although these 7 SISR methods could achieve a relatively good accuracy, what we should pay attention to is the gap between SR results and BICUBIC. These methods could improve the average accuracy 2.3$\%\sim$ 5.8$\%$, while ours could improve 10.7$\%\sim$14.6\%. We can also find that our TSRN could improve the accuracy on all of the three state-of-the-art recognizers. In the supplementary materials, we show the results of PSNR and SSIM and show that our TSRN could also surpass most of the state-of-the-art methods in PSNR and SSIM.

\begin{figure*}[ht]
    \centering
    \includegraphics[width=0.98\textwidth]{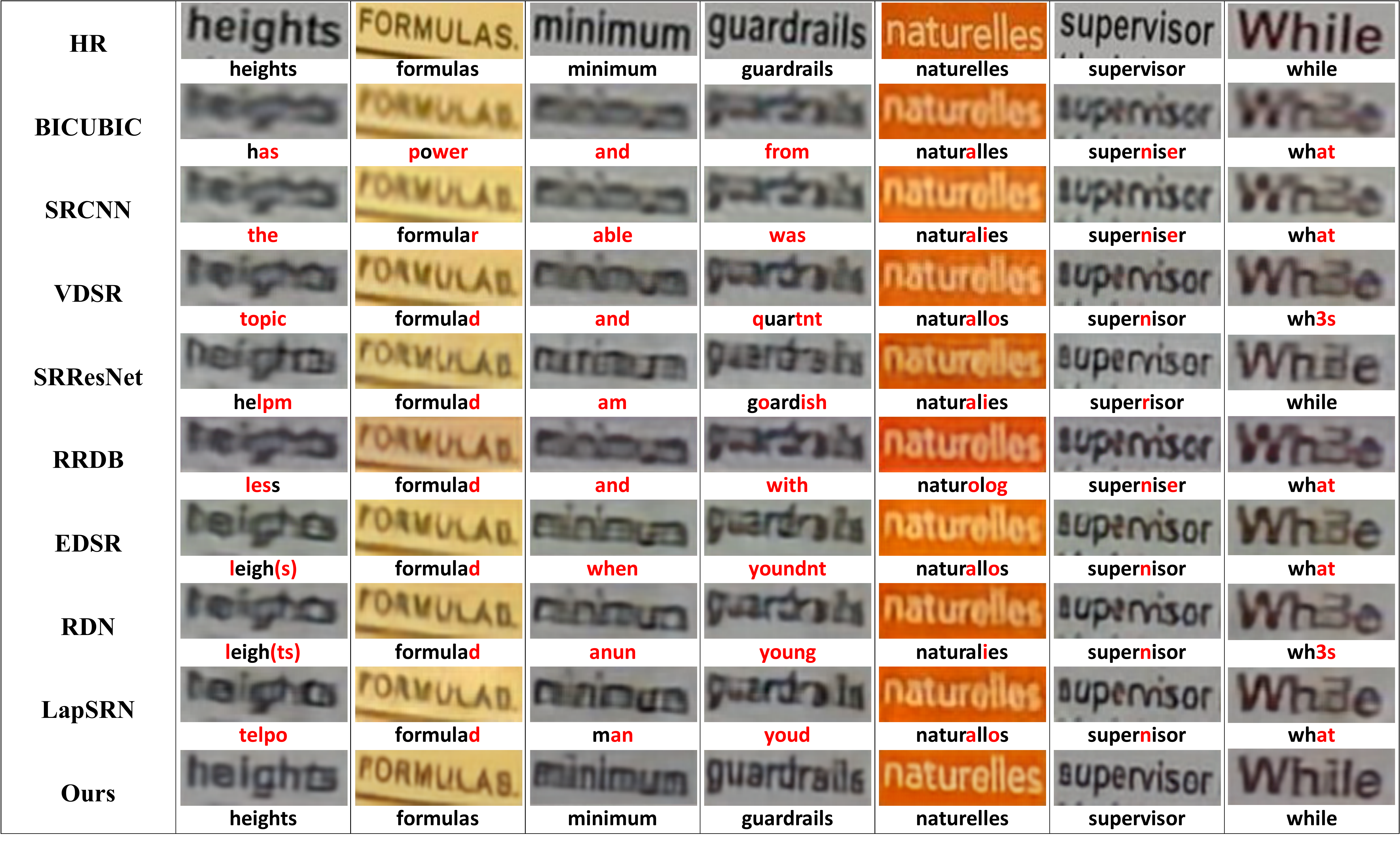}
    \caption{Visualization results of state-of-the-art SR methods on our proposed dataset TextZoom. The character strings under the images are recognition results of ASTER~\cite{aster}. Those in red denote wrong recognition.}
    \label{fig:display}
\end{figure*}

\section{Conclusion and Discussion}
In this work, we verify the importance of scene text image super-resolution task. We proposed the TextZoom dataset, which is, to the best of our knowledge, the first real paired scene text image super-resolution dataset. The TextZoom is well annotated and allocated and divided into three subset:easy, medium and hard. Through extensive experiments, we demonstrated the superiority of real data over synthetic data. To tackle text images super-resolution task, we build a new text-oriented SR method TSRN. Our TSRN clearly outperforms 7 SR methods. It also shows low-resolution text SR and recognition is far from being solved, thus more research effort is needed. 

In the future, we will capture more appropriately distributed text images. Extremely large and small images will be avoided. The images should also contain more kinds of languages, such as Chinese, French and Germany. 
We will also focus on new methods such as introducing recognition attention into the text super-resolution task.

\textbf{Acknowledge}
Xiang Bai was supported by the Program for HUST Academic Frontier Youth Team 2017QYTD08.
Chunhua Shen and his employer received no financial support for the research, authorship, and/or publication of this article.

\clearpage

\appendix
\centerline{\textbf{\Large{Appendix}}}

\section{Extensive Experiments on Our Method}

\subsection{Synthetic LR vs. TextZoom LR}
To demonstrate the superiority of paired scene text SR images, we compare the performance of the models trained on synthetic datasets and our TextZoom dataset. Traditional SISR tasks simply down-sample HR image by bicubic interpolation to generate corresponding LR images.
To illustrate the superiority of real LR over synthetic LR, we train our model on the bicubic down-sampled LR images and real LR images to show the performance.

\begin{table}[ht]
\begin{center}
\caption{The comparison of the models trained on synthetic LR and real LR. The listed results are the models evaluated on proposed TextZoom LR images. For better displaying, we calculated the average accuracy. The recognition accuracies are tested by the official released model of ASTER~\cite{aster}, MORAN~\cite{luo2019moran} and CRNN~\cite{crnn}. `Syn' denotes down-sampled LR and `Real' denotes proposed LR images.}
\vspace{-0.4cm}
\scalebox{0.65}{
    \input{tables/bd_rd2.tex}
    }

	\vspace{-0.3cm}
	\label{tab:bd_rd2}
\end{center}
\vspace{-0.5cm}
\end{table}

We selected SRResNet~\cite{srgan}, LapSRN~\cite{lapsrn} and our proposed method TSRN, and trained them on the synthetic LR and real LR datasets for a 2X model respectively. We trained 6 models in all and evaluated them on our proposed TextZoom subsets. From Table~\ref{tab:bd_rd2}, we can figure that the three methods trained on real LR (TextZoom) dataset outperform the models trained on synthetic LR obviously in accuracy. For our TSRN, the model trained on real LR could surpass the synthetic LR for nearly 9.0\% on ASTER and MORAN, and nearly 14.0\% on CRNN.

\subsection{Speed $\&$ Accuracy.}
In this task, we take the recognition accuracy as the most import evluation metric. To figure out whether it is wise to increase the accuracy at the cost of the extra computation consumption of TSRN, we compare the number of parameter, FLOPs and inference FPS of w and w/o super-resolution. The inference FPS means the FPS of recognizing the text images w or w/o SR. Through Table~\ref{tab:ablationspeed}, we can find that the proposed method is relatively tiny compared to the recognition network.  The FPS of `with TSRN' is nearly equal to direct recognition of attention based recognizer ASTER~\cite{aster} and MORAN~\cite{luo2019moran}. The FPS of CTC based recognizer CRNN decreases when adding the TSRN, but the improvement of accuracy is very considerable. So it would be a suitable manipulation to take super-resolution as a pre-processing procedure before recognition. (All of the FPSs were tested on a single GTX 1080Ti GPU with the same batch-size of 50.) 

\begin{table}[ht]
\center 
\caption{Computation and speed comparison between w or w/o super resolution when recognize TextZoom. `$\times$' means directly recognizing BICUBIC up-sampled LR images. `$\surd$' means recognizing after super-resolving images by our TSRN. \textbf{The inference FPS means the FPS of recognizing w or w/o SR.}}
\scalebox{0.75}{
    \input{tables/ablation4.tex}
    }
	
	\label{tab:ablationspeed}
\end{table}

\subsection{Binary Mask}

\begin{figure}[ht]
		\centering
		\setlength{\tabcolsep}{1.2mm}
		\scalebox{0.47}{
		\includegraphics[]{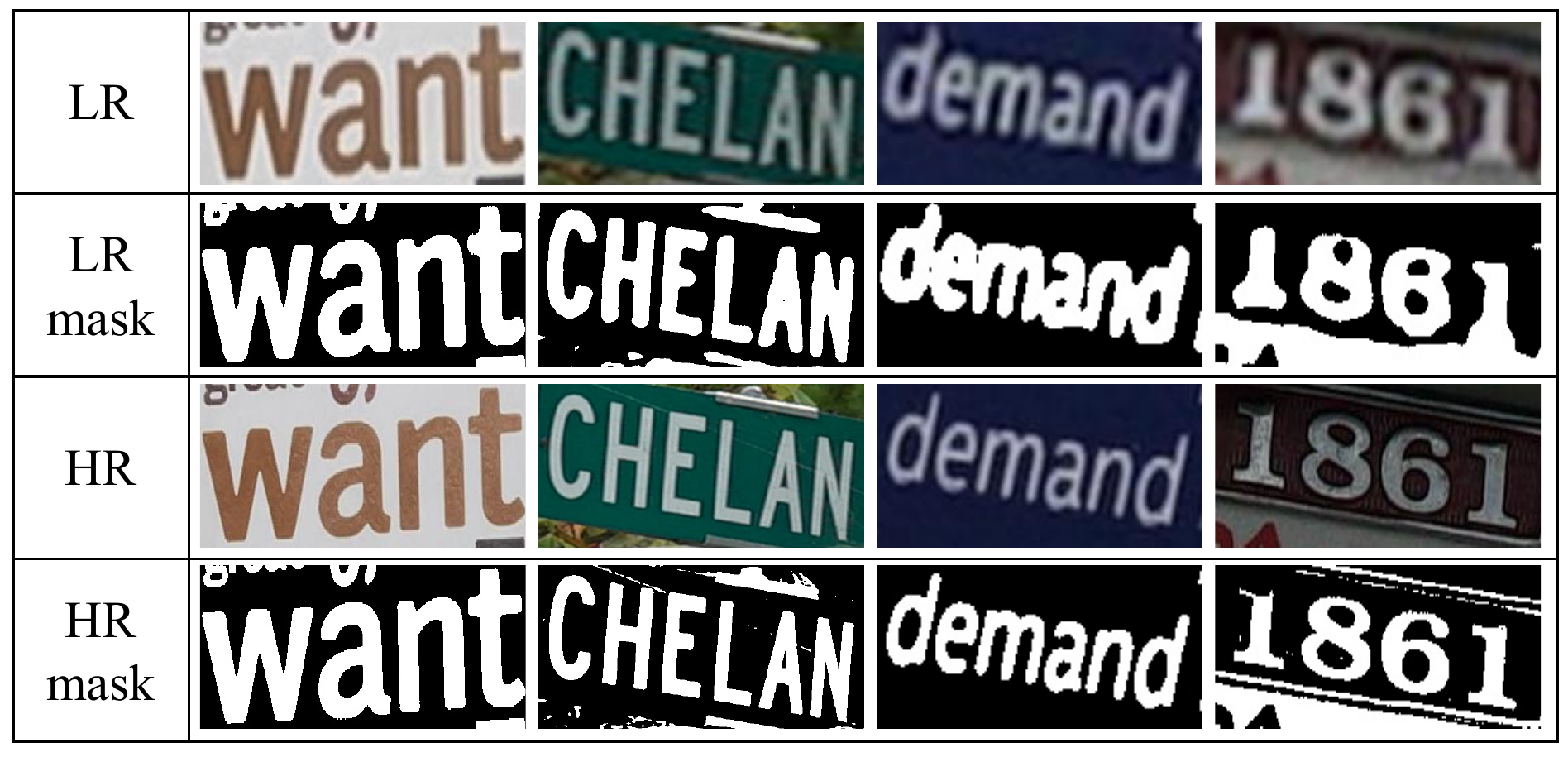}
		}
		\caption{The demonstration of the binary masks.}
		\label{fig:mask}
		\hspace{0.2cm}

	\end{figure}

In text images, the characters are usually in a unified color. The only texture information is the character color and background color. For brevity, we concatenate the binary mask with text images as input (Fig.~\ref{fig:mask}). The character regions render 1 and the background regions render 0. This input can be viewed as a transcendental semantic segmentation label of text images since most of the text images only contain 2 colors: the text color and background color. The masks are simply generated by calculating the average gray scale of the RGB images.

	\begin{table}[]
	    \centering
	    
	    \caption{The ablation study of binary masks.}
		{
		\input{./tables/ablation}
		}
	    \label{tab:my_label}
	\end{table}
	
\subsection{Discussion about SRB}
To build the best architecture of SRB, we gradually modify this two essential configuration: the number of hidden units and the number of blocks. Our method select 5 $\times$ SRB with 32 hidden units each. In this section, we do ablation study on this two component separately.

\textbf{1) Hidden Units.}
The BLSTMs are used to build sequence dependence in the text lines, so we hypothesize that more hidden units could get better performance.
By the experiments, we compare 0, 16, 32, 64, 128 of hidden layers. 0 Hidden Units represents SRResNet. The results demonstrate that the network would achieve best accuracy when the number of hidden unit equal 32 (Table~\ref{tab:hd}) Too many hidden units achieve lower performance since it already build the sequence-dependence well.

	\begin{table}[ht]
			\centering
		    \caption{Comparison between different number of hidden units of our proposed method on TextZoom.}
			{
			\input{./tables/ablation3}
			}
			\label{tab:hd}

	\end{table}
	
\begin{table}[]
    \centering
	\centering
    \caption{Comparison between different number of SRBs of our proposed method on TextZoom.}
    {
	\input{./tables/ablation2}
	}
\label{tab:srb_number}
  
\end{table}

\textbf{2) Block Number.}
To figure out whether we can achieve better performance by building deeper network, we stack different number of SRBs to compare the performance. In Table~\ref{tab:srb_number}, we compare our method with 4, 5, 6, 7 SRBs. We can find that more SRBs may not boost up the performance. The accuracy of 7 SRBs even decrease obviously. Stacking 5 SRBs, the network saturates and could get the best performance.

Our configuration of Sequence Residual Block is then shown in Table~\ref{tab:srb}. 

\begin{table}[ht]
\centering
\caption{Network configuration summary. The first row is the top layer. `k', `s' and `p' stand for kernel size, stride and padding size respectively.}
{
    \input{tables/srb.tex}
    }
	
	\label{tab:srb}
\end{table}

\subsection{PSNR $\&$ SSIM}
To calculate the PSNR[dB] and SSIM, we borrow the code from \url{https://github.com/open-mmlab/mmsr.} From Table~\ref{tab:psnr}, our PSNR of medium and hard subsets are not so good because PSNR is pixel-to-pixel calculated, while SSIM is calculated with a 11$\times$11 sliding kernel. The central alignment module would introduce slight pixel shift so the PSNR is somewhat lower than other SR methods. Usually, PSNR and SSIM could not represent the visual quality of the images~\cite{srgan}, in this task, it is also not so important compared to accuracy.

\begin{table}[ht]
    \centering
    \caption{PSNR and SSIM results of different SR methods on TextZoom.}
    \scalebox{0.9}{\input{./tables/psnr_ssim}}
    \label{tab:psnr}
\end{table}

\section{Central Alignment Module}
\begin{figure}
    \centering
    \includegraphics[width=1\textwidth]{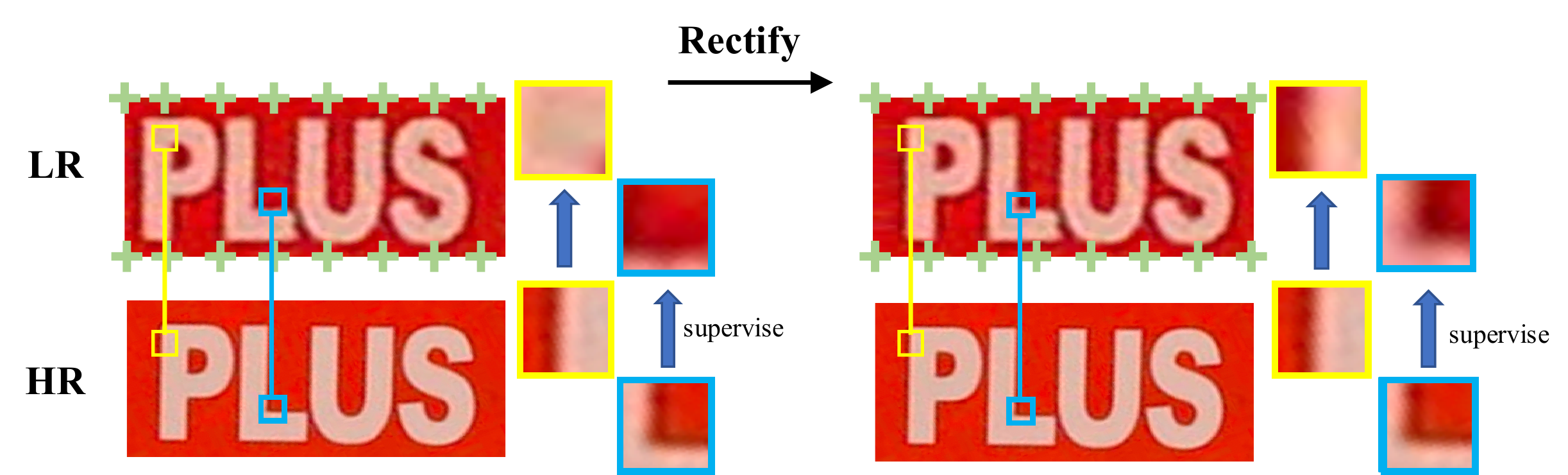}
    \caption{Demonstration of central alignment module.}
    \label{fig:stn}
\end{figure}

Our central alignment module is based on Spatial Transformation Network~\cite{stn}. The network predicts a set of control points and then then image is rectified by a Thin-Plate-Spline(TPS)~\cite{tps} transformation.
Our central alignment module mainly use horizontal or vertical shift. But sometimes the background region need different transformation scale to let the character region more central placed. So we use TPS transformation here to let the transformation flexible. As shown in Figure.~\ref{fig:stn}, the transformation is different between different points. 

\subsection{Performance on Manual Enlarged Misalignment}
We can find from ablation study that the central alignment module could improved the average accuracy for less than 2.0\%. Indeed, it can perform better on more misaligned text image pairs. To prove that, we do data augmentation aiming at generating more misaligned image pairs.
We crop our dataset TextZoom using a box with a 90\% width and 90\% height of the original image size randomly slide on the LR image, and get a region of 90\%$\times$90\% image. The HR images are not cropped. We train on the cropped dataset and evaluate on TextZoom.
In Table~\ref{tab:crop}, we show the performance of central alignment module on our manual cropped misalignment data.
From the results in Table.\ref{tab:crop}, we can find that the accuracy could be sharply improved. 

\begin{table}[ht]
    \centering
    \caption{Performance of w or w/o central alignment module on TextZoom which was trained on the mannual enlarged misaligned data.}
    \scalebox{1}{\input{./tables/crop}}
    \label{tab:crop}
\end{table}

\begin{figure}
    \centering
    \includegraphics[width=1\textwidth]{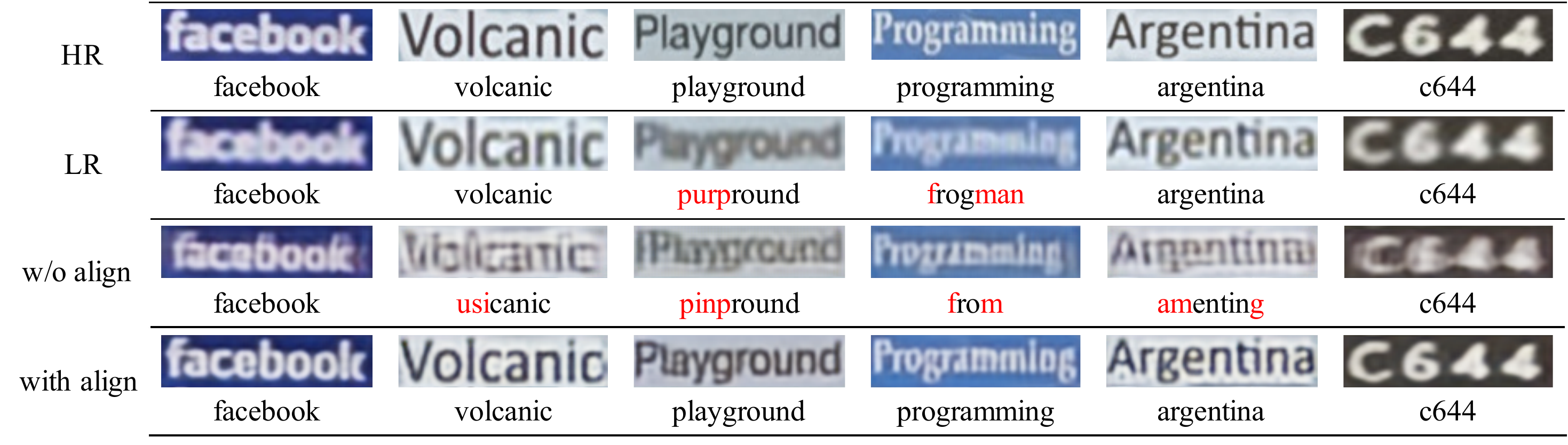}
 
    \caption{Comparison of w or w/o central align on enlarged misaligned data. The character strings under the images are the recognition results tested by ASTER~\cite{aster}. Those in red means wrongly recognition. For better display, we crop some obvious patches to compare the performance of w or w/o alignment.}
    \label{fig:vis_stn}
\end{figure}
We show visulization results in Fig.~\ref{fig:vis_stn}. The third row is the SR images trained without alignment. We can find that the double shadow and artifacts are very serve when trained without central alignment module. We can find that many words are still correctly recognized even with strong double shadow.

\subsection{Plugged into Other SR methods}
In this study, we compare the performance of w or w/o central alignment module on our dataset. (Table~\ref{tab:ablationstn}). We display the performance of six models: w and w/o Central Alignment Module on SRResNet, LapSRN and ours seperately. The improvement of central align on these methods illustrate that it is a conveniently pluggable module for SR networks, and all the performance could be improved.

\begin{table}[ht]
    \centering
    \caption{Comparison between w or w/o Central Alignment Module on TextZoom.}
   {\input{./tables/ablation6}}
    \label{tab:ablationstn}
\end{table}

\subsection{Comparison with CoBi Loss.}
CoBi Loss was proposed in~\cite{srraw} to tackle the misalignment. It is based on Contextual Loss~\cite{mechrez2018contextual}. It modiﬁed the nearest neighbor search and considers local contextual similarities with weighted spatial awareness. The CoBi Loss used pre-trained VGG-19 features and select several conv layers as deep features. Its formulation is shown in Eqn.~\ref{eqn:cobi}~\ref{eqn:cobi2}~\ref{eqn:cobi3}. The results are shown in Table~\ref{tab:cobi}. It is less practical in this task because the pre-trained model is trained on a classification dataset.
\begin{equation}
    CoBi(P,Q)=\frac{1}{N}\sum_{i}^{N}\mathop{min}_{j=1,...M}(\mathbb{D}_{p_{i},q_{j}}+\omega_{s}\mathbb{D}^{'}_{p_{i},q_{j}})
    \label{eqn:cobi}
\end{equation}

\begin{equation}
    \mathbb{D}^{'}_{p_{i},q_{j}}=||(x_{i},y_{i})-(x_{i},y_{i})||_{2}
    \label{eqn:cobi2}
\end{equation}

\begin{equation}
    CoBi Loss=CoBi_{RGB}(P,Q,n)+\lambda CoBi_{VGG}(P,Q)
    \label{eqn:cobi3}
\end{equation}

\begin{table}[ht]
    \centering
    \caption{Comparison between CoBi Loss and central alignment module.}
    {\input{./tables/cobi}}
    \label{tab:cobi}
\end{table}

\section{Detailed Information of TextZoom.}
\subsection{Annotation of SR-RAW and RealSR.}

\begin{figure}[ht]
\centering
\includegraphics[width=1\textwidth]{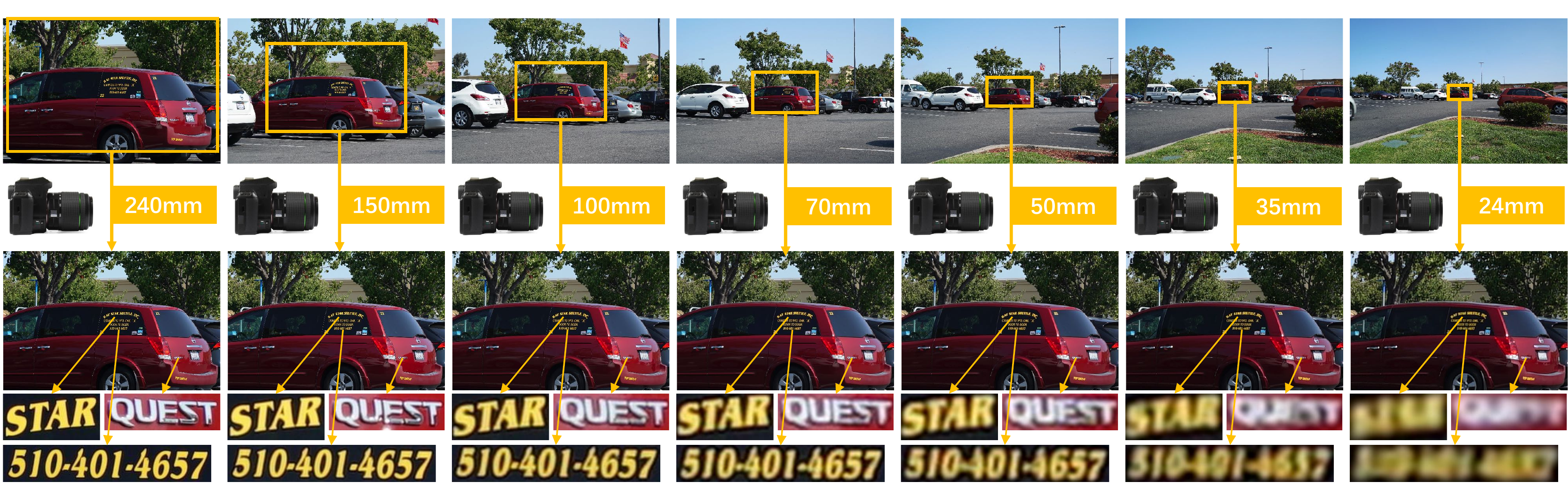}
\caption{The demonstration of the SR-RAW paired images and how we cropped text images. }
\label{fig:srraw}
\end{figure}

\textbf{SR-RAW}~\cite{srraw} is collected by seven different focal lengths with SONY FE camera, ranging from 24-240mm. We demonstrate it in Fig.~\ref{fig:srraw}.  There are totally 500 images in SR-RAW dataset, where 450 in train set and 50 in test set. The images are then aligned via field of view (FOV) matching and geometric transformation. The images captured in shorted focal lengths could be used as LR images while those captured in longer lengths as corresponding ground-truth. The author of SR-RAW~\cite{srraw} applied down-sample operation as offset when the ratio does not match precisely. For example, when use (35mm, 150mm) pairs to train a 4X model, the 150mm images should be down-sampled to 140mm at first. In our project, we follow this strategy in our dataset pre-processing. We annotate all the images taken from 240mm focal length which contains recognizable text in SR-RAW dataset. AS showed in Fig.~\ref{fig:srraw}, the focal length decreases from left to right, from 240mm to 24mm. The smaller the focal length, the smaller the field of view. The annotated text images have the same text contexts but different resolutions. We display three groups in Fig.~\ref{fig:srraw}: `STAR', `QUEST', `510-401-4657'. In this image, the text images cropped from 35mm and 24mm are hardly recognizable. How many clear images in a group of 7 images mainly depends on the height of original box in the 240mm focal lengths images.

In Table~\ref{tab:srraw}, we show the information of the cropped text images in SR-RAW. In the original images, Some groups of images do not have the 7th image, so the number of 24mm is less than the others. Through the table we can figure out that the recognition accuracy decreases obviously as the resolution degrades. We use the released ASTER~\cite{aster} model to test the accuracy.

\begin{table}[ht]
\centering
	\caption{The detailed information of the text images cropped from SR-RAW dataset. The 2nd to 7th groups of text images are cropped following the annotated bounding box in the 1st group.}
\scalebox{0.9}{
    \input{tables/srraw.tex}
    }

	\label{tab:srraw}
\end{table}

\begin{figure}[ht]
			\centering

			\scalebox{0.3}{
			\includegraphics[]{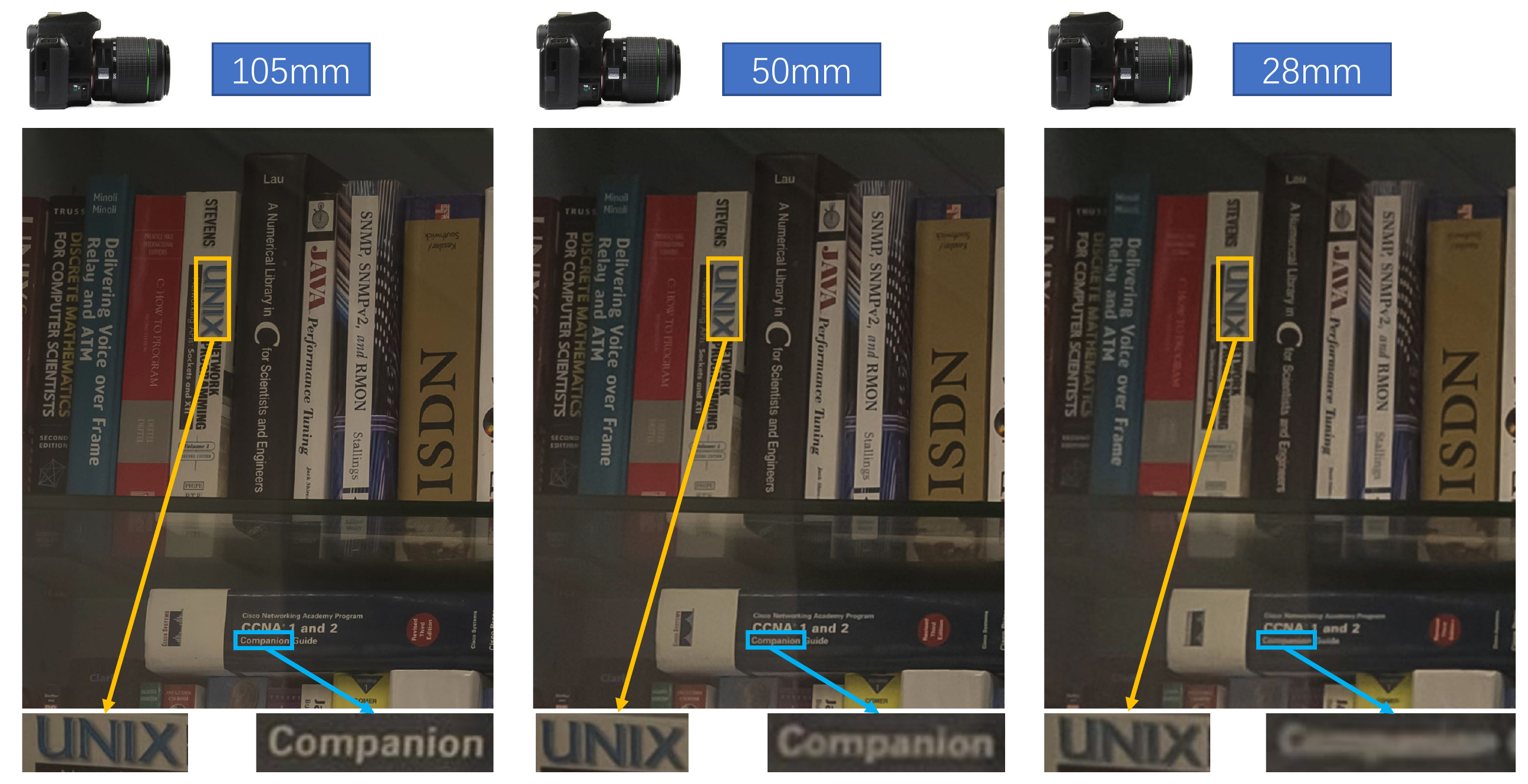}
			}
			\caption{The demonstration of the strategy we annotated the RealSR.}
			\label{fig:realsr}

	\end{figure}

\textbf{RealSR}~\cite{realsr} is captured by two Digital Single Lens Reflex(DSLR) cameras: Canon 5D3 and Nikon D810 with four focal lengths: 105mm, 50mm, 35mm, and 28mm. In RealSR~\cite{realsr}, the images taken by 105mm focal length are used to generate HR images, while images taken by 50mm, 35mm, 28mm are used to generate 2X, 3X, 4X LR images separately. separately. 
For convenience, we only crop the 105mm, 50mm and 28mm. The non-horizontal text images are rotated to the most suitable angle for recognition (see Fig.~\ref{fig:realsr}).
	
In Table~\ref{tab:realsr}, we briefly show the statistics of text images in RealSR.

	\begin{table}[]
	    \centering

		    \caption{The detailed information of the text images cropped from RealSR dataset.}
			\scalebox{1}{
			\input{./tables/realsr}
			}

			\label{tab:realsr}
	\end{table}
	
\textbf{Align.}
In RealSR, the author aligned the image pairs by introducing a pixel-wise registration algorithm which take  luminance difference into consideration. 
In SR-RAW~\cite{srraw}, the Euclidean motion model is used as the pre-processing procedure. During training, a contextual bilateral loss is proposed to leverage the misalignment, but a pre-trained model is needed, and it brings high computation consumption.
 We adapted their proposed pre-processing method to align the original images and cropped our dataset following our annotation principal. While in training, we used central alignment module as replacement.

\textbf{Accuracy by Height.}
The size of the cropped text boxes is diversed, 
We can figure that with the similar focal length, the accuracy of text images in RealSR is much higher than that in SR-RAW (Table~\ref{tab:realsr}~\ref{tab:srraw}). This mainly due to that the SR-RAW images are taken from a longer distance. So it is suitable to allocate images cropped from RealSR as subset~\textbf{easy}.

We divided the previous cropped images by height and found that the accuracy is relatively good when the height reaches 16-32 pixels, which is showed in Table~\ref{tab:srraw_divide}.  The images sized in (16-32) and (8-16) claim the majority in all the groups.

 The accuracy of the images smaller than 8 pixels are too low, which hardly have any value for restoration. The images are hardly recognizable, so we discard the images the height of which is less than 8 pixels. (8-16, 16-32) should be a good pair to form a 2X train set for STR super-resolution task. For example, the text images taken from 150mm focal length and height sized in 16-32 pixels would be taken as a ground-truth for the 70mm counterpart. So we selected all the images the height of which range from 16 pixels to 32 pixels as our ground-truth image and up-sample them to the size of 128$\times$32~(width$\times$height), and the corresponding 2X LR images to the size of 64$\times$16~(width$\times$height).

\begin{table*}[ht]
\centering
\caption{The recognition accuracy of the text images divided by height.}
\scalebox{1}{
    \input{tables/divided_acc.tex}
    }
	
	\label{tab:srraw_divide}
\end{table*}

\subsection{Statistical information.}
\begin{figure}[ht]
\centering
\scalebox{0.95}{\subfigure[Character distribution. ]{\includegraphics[width=0.45\textwidth]{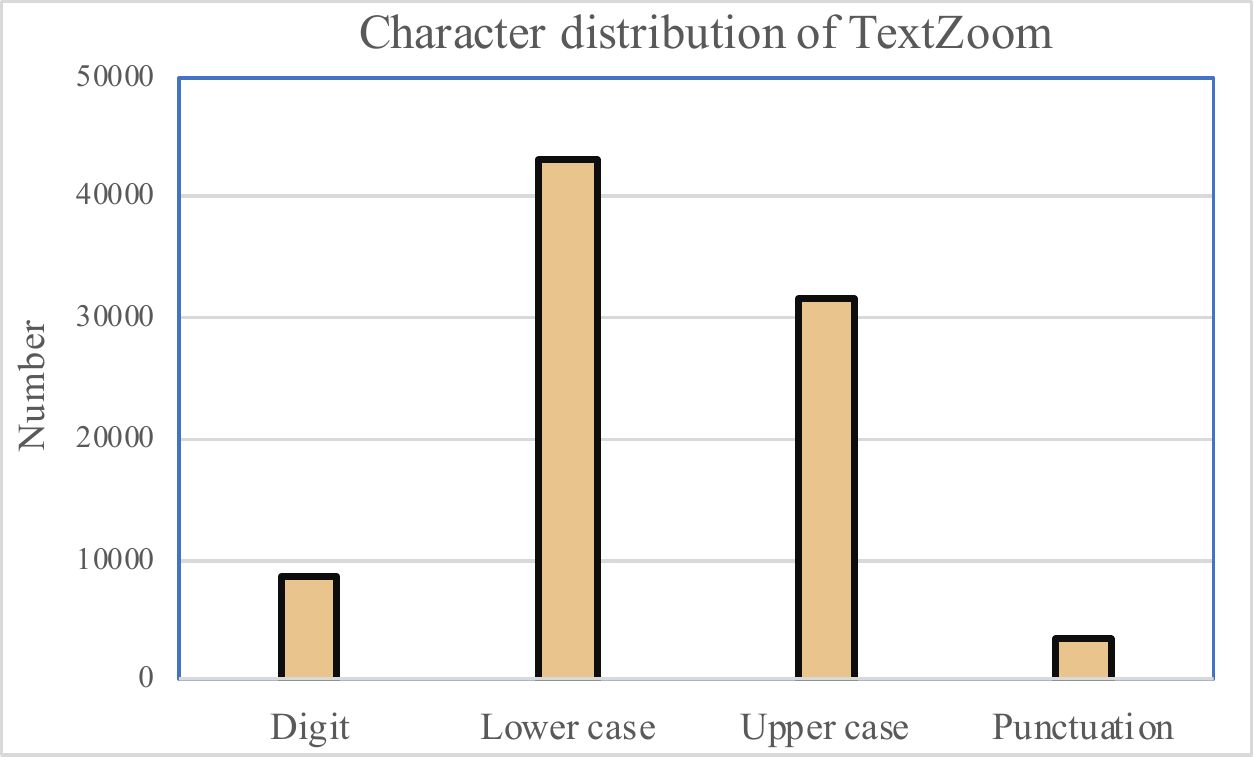}}}
\hspace{0.4cm}
\scalebox{0.95}{\subfigure[Character number in each image.]{\includegraphics[width=0.45\textwidth]{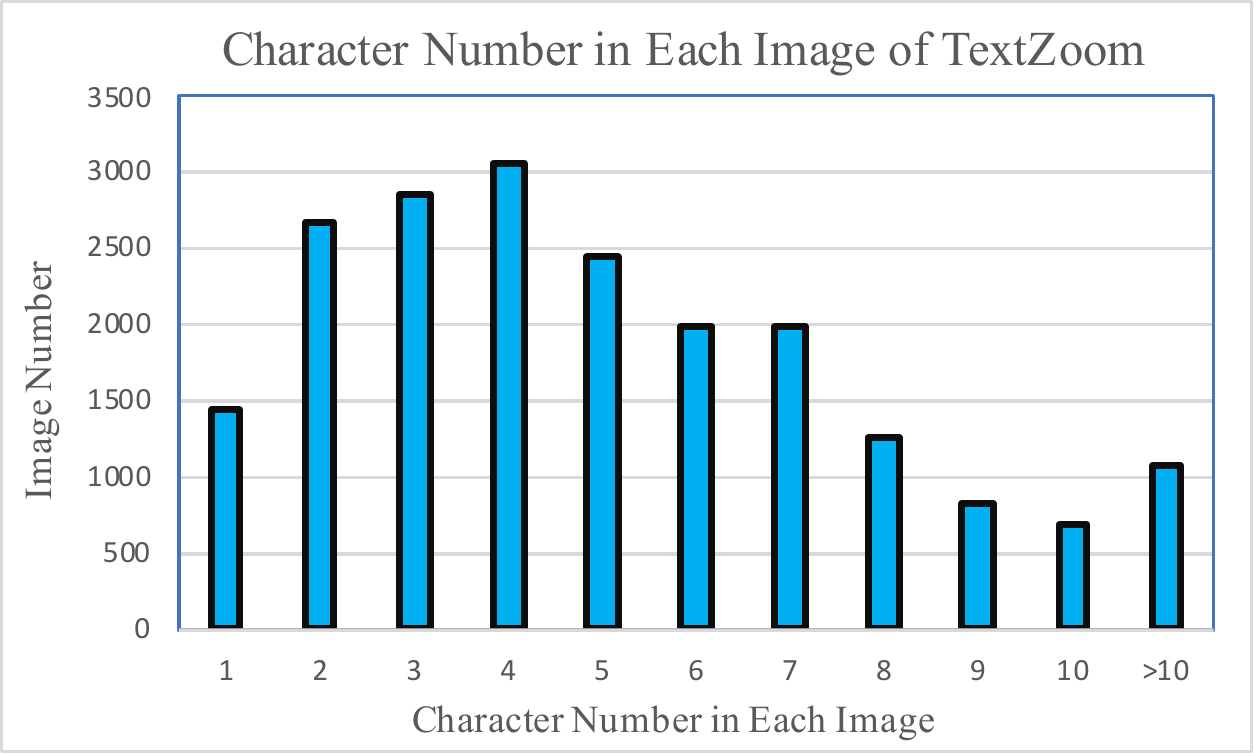}}}
\scalebox{0.95}{\subfigure[Direction of the bounding boxes. For better display, we list the logarithm of the ordinate number.]{\includegraphics[width=0.45\textwidth]{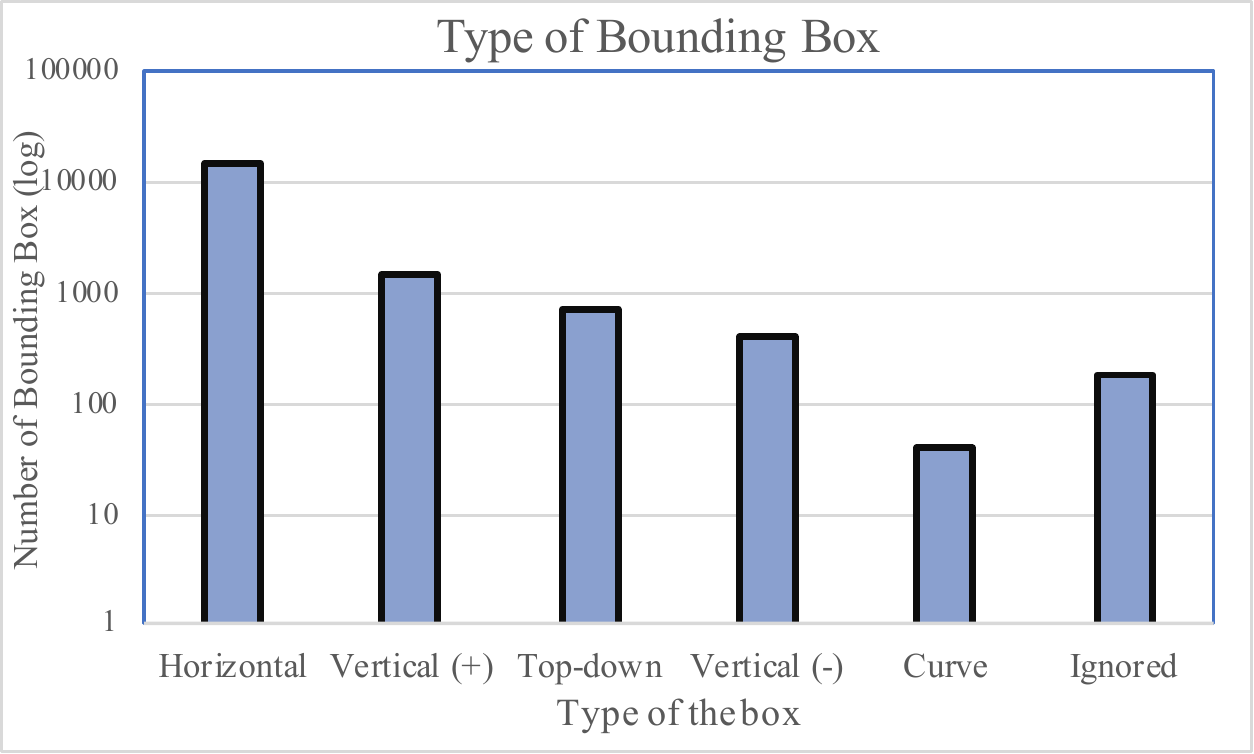}}}
\hspace{0.4cm}
\scalebox{0.95}{\subfigure[Distribution of text contents in TextZoom.]{\includegraphics[width=0.45\textwidth]{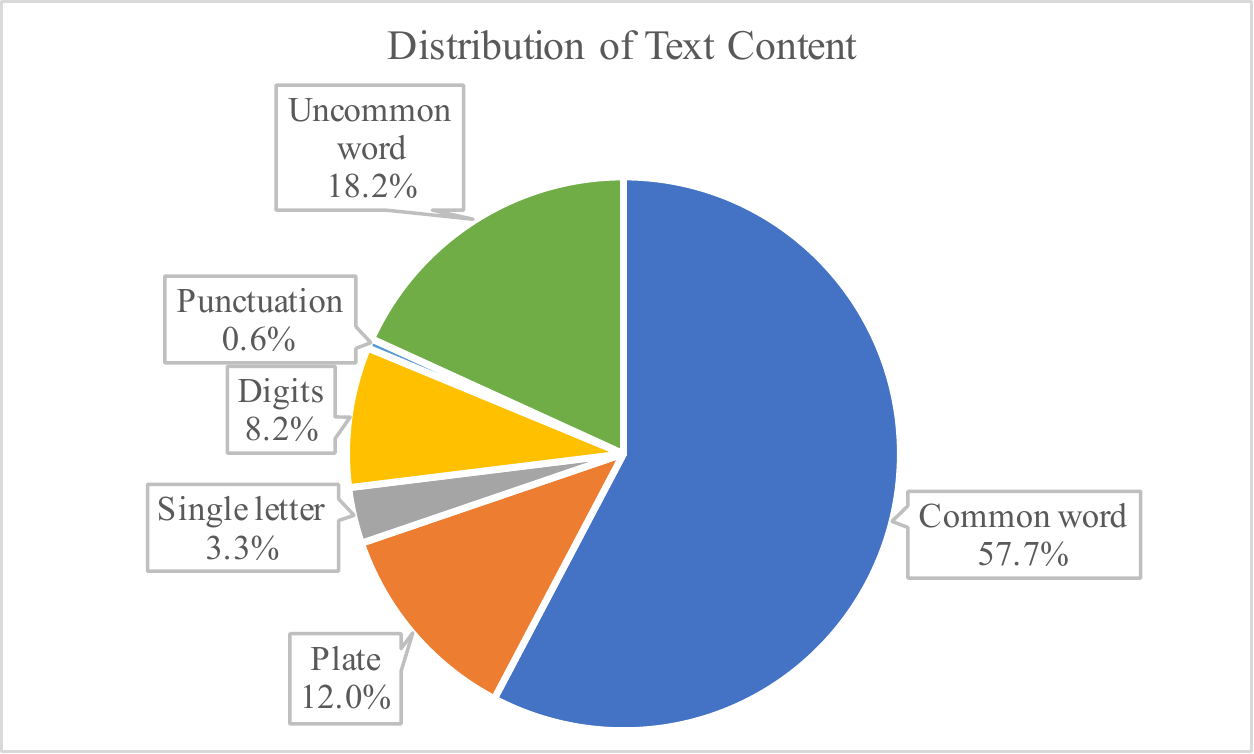}}}
\caption{Statistical information of TextZoom. }
\label{fig:stat}
\end{figure}

We display some useful statistical information in Fig.~\ref{fig:stat}. (a) Our dataset contains abundant characters and digits, including some punctuation. (b) Most of the lengths of the words range from 1-8 characters. (c) There are many randomly-placed boxes and books in the original images, so we count the direction type of the bounding boxes we annotated. `Horizontal' means that the text image is horizontal placed, easy to read. `Vertical(+)' denotes the text image is vertical and it should be rotated following the clockwise direction for 90 degrees, while `Vertical(-)' denotes following anti-clockwise direction for 90 degrees. `Top-down' denotes that the text image should be rotated 180 degrees for the best recognition. `Curve' denote the text image is curved. `Ignored' means that the text is illegal (not digits, English letters or punctuation). (d) Via the generic lexicon which has 90k common words used in ICDAR2015~\cite{icdar2015}, we figure that 57.5\% of the text contents are common English words. Plate includes car license plates, door number plates or street signs. They are the combination of digits, punctuation and letters. This kind of text account for 12\% because there are many street views in the original images. Uncommon word claims 18.2\% in all the texts. This kind of text are mainly rare words, phrases or compound words. Other meaningless strings like punctuation, single letter and digits account for the rest.

\subsection{Task Analysis}
\begin{figure}[ht]
\centering
\subfigure[Example images of easy subset.]{\includegraphics[width=1\textwidth]{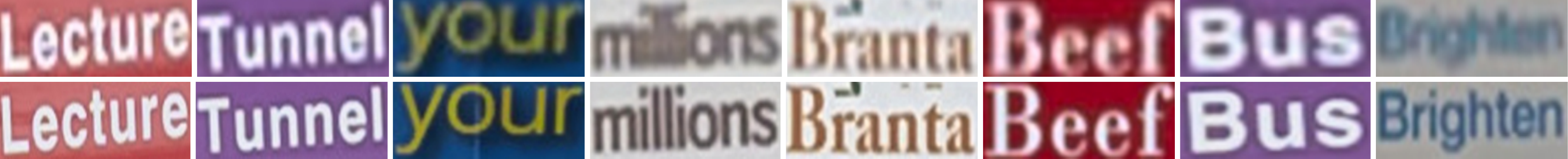}}
\subfigure[Example images of medium subset.]{\includegraphics[width=1\textwidth]{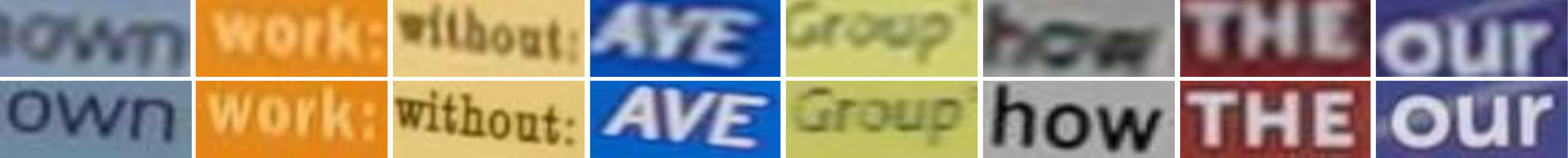}}
\subfigure[Example images of hard subset.]{\includegraphics[width=1\textwidth]{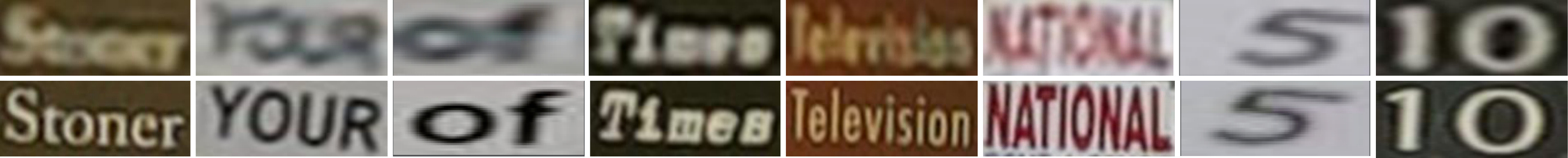}}
\caption{Demonstration of the images in TextZoom.  The misalignment and ambiguity becomes server as the difficulty increases.} 
\label{fig:misalign}
\end{figure}

Our dataset is challenging mainly for two reasons: the misalignment and ambiguity. 
Misalignment is unavoidable during data capture when the lens zoom in and out. Any slight camera movement could cause tens of pixels shift, especially the short focal lengths. And the pre-processing procedure cannot totally eliminate misalignment.
We display some example images in Fig.~\ref{fig:misalign}.

From Fig.~\ref{fig:misalign}, we can figure that the misalignment varies and no specific regulation can be found since we do not have pixel-level annotation of the word location. The three different subsets are allocated appropriately by the difficulty. The misalignment and ambiguity becomes server as the difficulty increases.
Note that the characters in HR images tend to locate in the center compared to those in LR. 
This mainly owing to that when we annotated the HR images, we artificially keep the text boxes at the centre of the images. 

\clearpage
\bibliographystyle{splncs04}
\bibliography{egbib}
\end{document}

%% file: tables/textzoom_allocation.tex
\begin{tabular}{p{80pt}<{\centering}|p{40pt}<{\centering}|p{40pt}<{\centering}p{40pt}<{\centering}p{40pt}<{\centering}}
\hline
\multirow{2}{*}{TextZoom}  & \multirow{2}{*}{train} & \multicolumn{3}{c}{\rule{0pt}{12pt}test} \\ \cline{3-5} 
                           &                \rule{0pt}{12pt} & easy    & medium & hard   \\ \hline
\rule{0pt}{12pt}Image number               & 17367                  & 1619    & 1411   & 1343   \\ \hline
\rule{0pt}{12pt}Accuracy(LR) & 35.7\%                 & 62.4\%  & 42.7\% & 31.6\% \\ \hline
\rule{0pt}{12pt}Accuracy(HR) & 81.2\%                 & 94.2\%  & 87.7\% & 76.2\% \\ \hline
\rule{0pt}{12pt}Gap & 45.5\%                 & 31.8\%  & 45.0\% & 44.6\% \\ \hline
\end{tabular}

%% file: tables/necessity.tex
\begin{tabular}{|p{70pt}<\centering|p{70pt}<\centering|p{70pt}<\centering|}
\hline
\multirow{2}{*}{Method} & \multicolumn{2}{c|}{\rule{0pt}{14pt}Recognition Accuracy}      \\ \cline{2-3} 
                        & \rule{0pt}{14pt} \textbf{TextZoom}   & \textbf{CommonLR}  \\ \hline
\rule{0pt}{14pt}Released                  & 47.2\%  & 75.3\% \\ \hline

\rule{0pt}{14pt}ReIm           &  52.6\%  & 79.3\%\\ \hline

\rule{0pt}{14pt}Fine-tune    & \textbf{59.3\%} & 73.2\% \\ \hline

\rule{0pt}{14pt}Ours          &   58.3\%    &   \textbf{80.3\%}  \\ \hline

\end{tabular}

%% file: tables/ablation_simple.tex
\begin{tabular}{p{20pt}<{\centering}p{130pt}<{\centering}|p{80pt}<{\centering}|p{40pt}<{\centering}|p{40pt}<{\centering}|p{40pt}<{\centering}|p{45pt}<{\centering}}
\hline
\multicolumn{3}{c|}{\rule{0pt}{12pt}Configuration}            & \multicolumn{4}{c}{Accuracy of ASTER~\cite{aster}} \\ \hline
\multicolumn{2}{c}{\rule{0pt}{12pt}Method }                       & Loss function  & easy    & medium    & hard   & \textbf{average}  \\ \hline
\multicolumn{1}{c|}{\rule{0pt}{12pt}~~~0~~~} &SRResNet                       & $L_{2}$ + $L_{tv}$ +$L_{p}$      &  69.6\%         &    47.6\% &    34.3\%  &    51.3\%  \\ 
\multicolumn{1}{c|}{\rule{0pt}{12pt}~~~1~~~} & 5$\times$SRBs               & $L_{2}$         &    74.5\%     &  53.3\%         &37.3\%    &  56.2\%    \\ 
\multicolumn{1}{c|}{\rule{0pt}{12pt}~~~2~~~} & 5$\times$SRBs + align        & $L_{2}$        &    74.8\%     &         55.7\%  &  39.6\%    &   57.8\%     \\ 
\multicolumn{1}{c|}{\rule{0pt}{12pt}~~~3~~~}& 5$\times$SRBs + align \textbf{(Ours)} & $L_{2} + L_{GP}$       &   \textbf{75.1\%}        &  \textbf{56.3\%}    & \textbf{40.1\%} &     \textbf{58.3\%}   \\ \hline
\end{tabular}

%% file: tables/main.tex
\begin{tabular}{p{60pt}<{\centering}|p{60pt}<{\centering}|p{30pt}<{\centering}p{30pt}<{\centering}p{30pt}<{\centering}p{38pt}<{\centering}|p{30pt}<{\centering}p{30pt}<{\centering}p{30pt}<{\centering}p{38pt}<{\centering}|p{30pt}<{\centering}p{30pt}<{\centering}p{30pt}<{\centering}p{38pt}<{\centering}}
\hline
\rule{0pt}{15pt}\multirow{2}{*}{Method} & \multirow{2}{*}{Loss Function} & \multicolumn{4}{c|}{Accuracy of ASTER~\cite{aster}}  & \multicolumn{4}{c|}{Accuracy of MORAN~\cite{luo2019moran}} & \multicolumn{4}{c}{Accuracy of CRNN~\cite{crnn}}  \\ \cline{3-14} 
                       \rule{0pt}{15pt} &                                & easy & medium & hard & \textbf{average} & easy   & medium   & hard  & \textbf{average} & easy   & medium   & hard & \textbf{average} \\ \hline

\rule{0pt}{15pt}BICUBIC                 &         $-$                       &      64.7\% &      42.4\%     &    31.2\%    & 47.2\%
& 60.6\%   &    37.9\%    & 30.8\%   &    44.1\%  &   36.4\%  & 21.1\%  & 21.1\% & 26.8\%\\ \hline
\rule{0pt}{15pt}SRCNN~\cite{srcnn}                   & $L_{2}$                             &  69.4\%       &     43.4\%      &      32.2\%   & 49.5\%  &   63.2\% &     39.0\%   &  30.2\%  &    45.3\%  &  38.7\%   &  21.6\% & 20.9\% & 27.7\% \\ \hline
\rule{0pt}{15pt}VDSR~\cite{vdsr}            & $L_{2}$                            &     71.7\%     &   43.5\%        &   34.0\%    & 51.0\%     &    62.3\%    & 42.5\%   &   30.5\% &  46.1\% &    41.2\% &   25.6\% & 23.3\%& 30.7\%  \\ \hline
\rule{0pt}{15pt}SRResNet~\cite{srgan}                & $L_{2}+L_{tv}+L_{p}$              &    69.6\%     &       47.6\%    &     34.3\%  & 51.3\% &       60.7\%     &  42.9\%  &  32.6\%   &  46.3\% &  39.7\%     &  27.6\%  &  22.7\% & 30.6\%  \\ \hline
\rule{0pt}{15pt}RRDB~\cite{esrgan}                    & $L_{1}$                           &   70.9\%      &      44.4\%     &   32.5\%      & 50.6\% &       63.9\%    & 41.0\%   & 30.8\%  &  46.3\% & 40.6\%    & 22.1\%   &21.9\% & 28.9\%\\ \hline
\rule{0pt}{15pt}EDSR~\cite{edsr}                    & $L_{1}$                             &    72.3\%     &     48.6\%      &     34.3\%    & 53.0\% &      63.6\%    &  45.4\%  &   32.2\%   &  48.1\%   &42.7\% &  29.3\% &24.1\% &  32.7\% \\ \hline
\rule{0pt}{15pt}RDN~\cite{rdn}                    & $L_{1}$                             &     70.0\%    &      47.0\%     &    34.0\%     & 51.5\%  &      61.7\%      &  42.0\%  &  31.6\%  &  46.1\%    &   41.6\%  &  24.4\%  &    23.5\% & 30.5\% \\ \hline
\rule{0pt}{15pt}LapSRN~\cite{lapsrn}                  & $Charbonnier$                &   71.5\%    &         48.6\%     &     35.2\%    & 53.0\%     &   64.6\%     & 44.9\% & 32.2\% &  48.3\% & 46.1\%   &   27.9\% & 23.6\% &  33.3\% \\ \hline
                 
\rule{0pt}{15pt}\textbf{TSRN(ours)}              & $L_{2}+L_{GP}$                    &   \textbf{75.1\%}
& \textbf{56.3}\% & \textbf{40.1\%} & \textbf{58.3\%} &  \textbf{70.1}\%  & \textbf{53.3\%}       &  \textbf{37.9\%}  &   \textbf{54.8\%}   & \textbf{52.5\%}    & \textbf{38.2\%}  & \textbf{31.4\%} &
 \textbf{41.4\%} \\ \hline\hline
 \multicolumn{2}{c|}{\rule{0pt}{15pt}\textbf{Improvement of TSRN}}                          &   \textbf{10.4\%}
& \textbf{13.9\%} & \textbf{8.9\%} & \textbf{11.1\%} &  \textbf{9.5\%}  & \textbf{15.4\%}       &  \textbf{7.1\%}  &   \textbf{10.7\%}   & \textbf{16.1\%}    & \textbf{17.1\%}  & \textbf{10.3\%} &
 \textbf{14.6\%} \\ \hline
\end{tabular}

%% file: tables/bd_rd2.tex
\begin{tabular}{|p{50pt}<{\centering}|p{45pt}<{\centering}|p{31pt}<{\centering}p{31pt}<{\centering}p{31pt}<{\centering}p{38pt}<{\centering}|p{31pt}<{\centering}p{31pt}<{\centering}p{31pt}<{\centering}p{38pt}<{\centering}|p{31pt}<{\centering}p{31pt}<{\centering}p{31pt}<{\centering}p{38pt}<{\centering}|}
\hline
\rule{0pt}{14pt}\multirow{2}{*}{Method} & \multirow{2}{*}{train data} & \multicolumn{4}{c|}{Accuracy of ASTER~\cite{aster}}  & \multicolumn{4}{c|}{Accuracy of MORAN~\cite{luo2019moran}} & \multicolumn{4}{c|}{Accuracy of CRNN~\cite{crnn}}  \\ \cline{3-14} 
                       \rule{0pt}{14pt} &                                & easy & medium & hard & \textbf{average} & easy   & medium   & hard & \textbf{average} & easy   & medium   & hard & \textbf{average}  \\ \hline

\rule{0pt}{14pt}BICUBIC                 &         $-$                       &      64.7\% &      42.4\%     &    31.2\%    & 47.2\%& 60.6\%   &    37.9\%    & 30.8\%   &    44.1\%  &   36.4\%  & 21.1\%  & 21.1\% & 26.8\% \\ \hline
\multirow{2}{*}{SRResNet}              & \rule{0pt}{14pt}Syn            &    66.4\%   &     44.4\%    &    32.4\%  &  48.9\%  
&   \textbf{61.8\%}    &   39.6\%     & 31.0\%    &    45.2\%   &     37.4\%    &   21.6\%   &  21.2\%    &  27.3\%     \\ 
\rule{0pt}{14pt}               & Real                      &           \textbf{69.4\%}     &  \textbf{47.3\%}    &  \textbf{34.3\%}     &   \textbf{51.3\%}   &       60.7\%     &  \textbf{42.9\%}  &  \textbf{32.6\%}   &  \textbf{46.3\%} &  \textbf{39.7\%}     &  \textbf{27.6\%}  &  \textbf{22.7\%} & \textbf{30.6\%}  \\
\hline
\multirow{2}{*}{LapSRN}              & \rule{0pt}{14pt}Syn            &   66.5\%    &    43.9\%     &   32.2\%   &  48.7\%   &    61.8\%  &    39.0\%    &   30.7\%   &    44.9\%   &    37.5\%   &  21.8\%  &   20.9\% &    27.3\%    \\ 
\rule{0pt}{14pt}               & Real        &   \textbf{71.5\%}    &         \textbf{48.6\%}     &     \textbf{35.2\%}    & \textbf{53.0\%}     &   \textbf{64.6\%}     & \textbf{44.9\%} & \textbf{32.2\%} &  \textbf{48.3\%} & \textbf{46.1\%}   &   \textbf{27.9\%} &  \textbf{23.6\%} &  \textbf{33.3\%}   \\\hline

\multirow{2}{*}{TSRN(ours)}              & \rule{0pt}{14pt}Syn            &   67.5\%    &    45.3\%     &   33.0\%   &   49.7\%  &    61.7\%   &  40.4\%      &  30.6\%   &   45.3\%    &   37.8\%     &  22.0\%   &    21.0\% &   27.6\%   \\ 
\rule{0pt}{14pt}               & Real                    
&   \textbf{75.1\%}
& \textbf{56.3\%} & \textbf{40.1\%} & \textbf{58.3\%} &  \textbf{70.1\%}  & \textbf{53.3\%}       &  \textbf{37.9\%}  &   \textbf{54.8\%}   & \textbf{52.5\%}    & \textbf{38.2\%}  & \textbf{31.4\%}  &  \textbf{41.4\%} \\\hline
\end{tabular}

%% file: tables/ablation4.tex
\begin{tabular}{|p{60pt}<{\centering}|p{60pt}<{\centering}|p{80pt}<{\centering}|p{80pt}<{\centering}|p{80pt}<{\centering}|p{60pt}<{\centering}|}
\hline
\multicolumn{6}{|c|}{\rule{0pt}{12pt}\textbf{Computation Cost Analysis}}\\ \hline
\rule{0pt}{12pt}Recognizer             & TSRN(ours)& Average Accuracy  & FLOPs &Parameters    & \textbf{Inference FPS}   \\ \hline
\multirow{2}{*}{ASTER~\cite{aster}} & \rule{0pt}{12pt}$\times$   & 47.2\%  & 4.72G &20.99M             & 21.97 \\ 

 & \rule{0pt}{12pt}$\surd$  & 58.3\%  (+10.1\%)  & 4.72G + 0.72G &  20.99M + 2.8M  & 21.67 \\ \hline
                       
\multirow{2}{*}{MORAN~\cite{luo2019moran}} & \rule{0pt}{12pt}$\times$   & 44.1\% & 0.73G & 20.3M             & 63.2 \\ 
& \rule{0pt}{12pt}$\surd$   & 54.8\%  (+10.7\%) & 0.73G + 0.72G & 20.3M+2.8M    & 59.6 \\ \hline

\multirow{2}{*}{CRNN~\cite{crnn}} & \rule{0pt}{12pt}$\times$ & 26.8\%  & 0.64G & 8.3M       & 514.7 \\ 
    
    & \rule{0pt}{12pt}$\surd$ & 41.4\% (+14.6\%)  & 0.64G + 0.72G  & 8.3M + 2.8M  & 340.6 \\ \hline
\end{tabular}

%% file: tables/ablation.tex
\begin{tabular}{|p{55pt}<{\centering}|p{40pt}<{\centering}|p{40pt}<{\centering}|p{40pt}<{\centering}|p{40pt}<{\centering}|}
\hline
\multicolumn{5}{|c|}{\rule{0pt}{14pt}\textbf{Ablation Study of Masks}}                                   \\ \hline
\multirow{2}{*}{Configuration} & \multirow{2}{*}{Mask} & \multicolumn{3}{c|}{\rule{0pt}{14pt}Accuracy} \\ \cline{3-5} 
                        &                       & \rule{0pt}{14pt}easy & medium   & hard    \\ \hline
\multirow{2}{*}{5 $\times$ SRB}   &\rule{0pt}{14pt} $\times$                   &   73.9\% &      51.6\%    &     36.0\%     \\ \cline{2-5} 
                        &\rule{0pt}{14pt} $\surd$                     &      \textbf{74.5\%}   & \textbf{53.3}\%      &   \textbf{37.3\%}   \\ \hline
\end{tabular}

%% file: tables/ablation3.tex
\begin{tabular}{|p{40pt}<{\centering}|p{65pt}<{\centering}|p{40pt}<{\centering}|p{40pt}<{\centering}|p{40pt}<{\centering}|}
\hline
\multicolumn{5}{|c|}{\rule{0pt}{ 14pt}\textbf{Ablation Study of Hidden Units}}\\\hline
 \multicolumn{2}{|c|}{\rule{0pt}{ 14pt}Configuration}                           & \multicolumn{3}{c|}{Accuracy} \\ \cline{1-5}
                      \rule{0pt}{14pt}SRBs            & Hidden Units      &   easy   &  medium   &   hard                                    \\ \hline
 \multirow{5}{*}{5}    
  & \rule{0pt}{ 14pt}0          & 69.6\%      & 48.3\%    & 34.3\%    \\\cline{2-5}
                                 
&   \rule{0pt}{ 14pt}16            & 71.6\%           & 52.1\%    & 36.3\%          \\ \cline{2-5} 
                                                            & \rule{0pt}{ 14pt}\textbf{32} &          \textbf{74.5\%}      & \textbf{53.3\%}    & \textbf{37.3\%}  
                                                            \\ \cline{2-5}
     & \rule{0pt}{ 14pt}64         & 71.9\%         & 50.8\%    & 35.8\%      \\\cline{2-5}
            & \rule{0pt}{ 14pt}128                 &   71.4\%  &      47.3\%     &    33.1\%           \\ \hline

\end{tabular}

%% file: tables/ablation2.tex
\begin{tabular}{|p{40pt}<{\centering}|p{60pt}<{\centering}|p{40pt}<{\centering}|p{40pt}<{\centering}|p{40pt}<{\centering}|}
\hline
\multicolumn{5}{|c|}{\rule{0pt}{ 16pt}\textbf{Ablation Study of SRBs}}\\\hline
 \multicolumn{2}{|c|}{\rule{0pt}{ 16pt}Configuration}                           & \multicolumn{3}{c|}{Metrics}\\ \cline{1-5}
                  \rule{0pt}{16pt}SRBs               & Hidden Units   &  easy       & medium  &  hard                        \\ \hline
          
  \rule{0pt}{16pt}4               &  \multirow{4}{*}{32}  & 73.3\%  & 52.1\%    & 35.8\%                \\ 
                             \cline{1-1} \cline{3-5}

                   \rule{0pt}{16pt}\textbf{5}                &          & \textbf{74.5}\%       & \textbf{53.3\%}    & \textbf{37.3}\%               \\\cline{1-1} \cline{3-5}
\rule{0pt}{ 16pt}6                                  & & 74.1\% & 52.7\%    & 37.0\%       \\ \cline{1-1} \cline{3-5}
 \rule{0pt}{ 16pt}7                  &                   & 72.3\%   & 50.9\%    &35.6\%            \\ \hline
\end{tabular}

%% file: tables/srb.tex
\begin{tabular}{|p{100pt}<{\centering}|p{120pt}<{\centering}|}
\hline

\rule{0pt}{10pt}\textbf{Type}                 & \textbf{Configurations}             \\ \hline
\hline
\rule{0pt}{10pt}FeatureMap           & B$\times$64$\times$Height$\times$Wdith          \\ \hline
\rule{0pt}{10pt}Convolution          & \#maps:64, k:3$\times$3, s:1 p:1 \\ \hline
\rule{0pt}{10pt}BatchNormalization   &                            \\ \hline
\rule{0pt}{10pt}PReLU                &                            \\ \hline
\rule{0pt}{10pt}Convolution          & \#maps:64, k:3$\times$3, s:1 p:1 \\ \hline
\rule{0pt}{10pt}BatchNormalization   &                            \\ \hline
\rule{0pt}{10pt}Convolution          & \#maps:64, k:1$\times$1, s:1 p:0 \\ \hline
\rule{0pt}{10pt}Permutuation         &                            \\ \hline
\rule{0pt}{10pt}Bi-LSTM              & \#hidden\_units: 32        \\ \hline
\rule{0pt}{10pt}Map-to-Sequence      &                            \\ \hline
\rule{0pt}{10pt}Permutuation         &                            \\ \hline
\rule{0pt}{10pt}Bi-LSTM              & \#hidden\_units: 32        \\ \hline
\rule{0pt}{10pt}Map-to-Sequence      &                            \\ \hline
\rule{0pt}{10pt}Permutuation         &                            \\ \hline
\rule{0pt}{10pt}Short Cut Connection &                            \\ \hline
\rule{0pt}{10pt}FeatureMap           & B$\times$64$\times$Height$\times$Wdith          \\ \hline
\end{tabular}

%% file: tables/psnr_ssim.tex
\begin{tabular}{p{60pt}<{\centering}|p{60pt}<{\centering}|p{40pt}<{\centering}p{40pt}<{\centering}p{40pt}<{\centering}|p{40pt}<{\centering}p{40pt}<{\centering}p{40pt}<{\centering}}
\hline
\rule{0pt}{12pt}\multirow{2}{*}{Method} & \multirow{2}{*}{Loss Function} &  \multicolumn{3}{c|}{PSNR} & \multicolumn{3}{c}{SSIM}  \\ \cline{3-8} 
\rule{0pt}{12pt} &  &   easy   & medium   & hard  & easy   & medium   & hard  \\ \hline

\rule{0pt}{12pt}BICUBIC                 &         $-$                      &    22.35    &     18.98     &   19.39    &    0.7884    &      0.6254    &    0.6592  \\ \hline
\rule{0pt}{12pt}SRCNN~\cite{srcnn}                   & $L_{2}$                  &   23.48     & \textbf{19.06}    &   19.34    &     0.8379   &     0.6323     &     0.6791  \\ \hline
\rule{0pt}{12pt}VDSR~\cite{vdsr}            & $L_{2}$                   &  24.62  &    18.96    &   \textbf{19.79}    &    0.8631     &    0.6166     &    0.6989  \\ \hline
\rule{0pt}{12pt}SRResNet~\cite{srgan}                & $L_{2}+L_{tv}+L_{p}$          &   24.36     &          18.88 &    19.29   &    0.8681    &    0.6406     &   0.6911   \\ \hline
\rule{0pt}{12pt}RRDB~\cite{esrgan}                    & $L_{1}$                 &   22.12  &     18.35     &  19.15     &    0.8351    &     0.6194     &    0.6856  \\ \hline
\rule{0pt}{12pt}EDSR~\cite{edsr}                    & $L_{1}$               &  24.26    &      18.63   & 19.14      &    0.8633    &   0.6440     &  0.7108   \\ \hline
\rule{0pt}{12pt}RDN~\cite{rdn}                    & $L_{1}$                 & 22.27    &      18.95    &  19.70     &    0.8249    &     0.6427     &    0.7113 \\ \hline
\rule{0pt}{12pt}LapSRN~\cite{lapsrn}                  & $Charbonnier$         &24.58     &    18.85      &   19.77    &    0.8556    &     0.6480     &   0.7087   \\ \hline
                 
\rule{0pt}{12pt}\textbf{TSRN(ours)}              & $L_{2}+L_{GP}$        & \textbf{25.07} & 18.86 & 19.71 & \textbf{0.8897} & \textbf{0.6676} &\textbf{0.7302} 
 \\ \hline
\end{tabular}

%% file: tables/crop.tex
\begin{tabular}{|p{80pt}<{\centering}|p{50pt}<{\centering}|p{50pt}<{\centering}|p{50pt}<{\centering}|p{50pt}<{\centering}|}
\hline

 \multirow{2}{*}{Method} & \multicolumn{4}{c|}{\rule{0pt}{12pt}Accuracy}  \\ \cline{2-5} 
                               & \rule{0pt}{12pt}easy & medium & hard & average \\ \hline
  \rule{0pt}{12pt} 5$\times$SRB                      & 66.8\% & 50.0\%   & 35.0\% &      51.6\%   \\ 
     \rule{0pt}{12pt}          5$\times$SRB+Align     & 74.4\% & 55.6\%   & 38.8\%&    57.4\%     \\ 
\rule{0pt}{12pt}Improvement  & +7.6\%  & +5.6\% &  +3.8\% & +5.8\% \\\hline
\end{tabular}

%% file: tables/ablation6.tex
\begin{tabular}{|p{45pt}<{\centering}|p{50pt}<{\centering}|p{35pt}<{\centering}|p{35pt}<{\centering}|p{35pt}<{\centering}|p{35pt}<{\centering}|}
\hline
\multirow{2}{*}{Method}   & \multirow{2}{*}{Alignment} & \multicolumn{4}{c|}{\rule{0pt}{11pt}Accuracy} \\ \cline{3-6} 
                          &                                & \rule{0pt}{11pt}  easy & medium   & hard   & average \\ \hline\hline

\multirow{2}{*}{SRResNet} &\rule{0pt}{11pt} $\times$                      &  69.6\%          &     47.6\%       & 34.3\%    & 51.7\% \\ \cline{2-6} 
                           &\rule{0pt}{11pt} $\surd$  &   \textbf{70.0\%}                            &  \textbf{ 49.6\%}         & \textbf{36.0\%}     & \textbf{53.0\%}   \\ \hline
                          \hline
\multirow{2}{*}{LapSRN} &\rule{0pt}{11pt} $\times$                      & 71.5\%             &     48.6\%       & 35.2\%    & 53.0\%   \\ \cline{2-6} 
                          &\rule{0pt}{11pt} $\surd$                 &   \textbf{71.7\%}             &    \textbf{50.3\%}        &   \textbf{35.7\%}   &  \textbf{53.7\%} \\ \hline
                          \hline

\multirow{2}{*}{5 $\times$ SRB}     & \rule{0pt}{11pt}$\times$                       &  74.5\%         &      53.3\%      & 37.3\%       & 56.2\%   \\ \cline{2-6} 
                          &\rule{0pt}{11pt} $\surd$        & \textbf{74.8\%}                   & \textbf{55.7\%}           &     \textbf{39.6\%}    &   \textbf{57.8\%}   \\ 
                          
                          \hline
\end{tabular}

%% file: tables/cobi.tex
\begin{tabular}{|p{100pt}<{\centering}|p{40pt}<{\centering}|p{40pt}<{\centering}|p{40pt}<{\centering}|}
\hline
\multirow{2}{*}{Method} &\multicolumn{3}{c|}{\rule{0pt}{12pt}Accuracy} \\ \cline{2-4} 
                            & \rule{0pt}{12pt}easy    & medium    & hard    \\ \hline
\rule{0pt}{12pt}  CoBi Loss                      &    74.0\%     &     51.6\%     &   36.0\%      \\ \hline
 \rule{0pt}{12pt}$L_{2}$  + alignment                           &     74.8\%    &      55.7\%     &   39.6\%      \\ \hline
\end{tabular}

%% file: tables/srraw.tex
    \begin{tabular}{p{100pt}<{\centering}|p{36pt}<{\centering}p{36pt}<{\centering}p{36pt}<{\centering}p{36pt}<{\centering}p{36pt}<{\centering}p{36pt}<{\centering}p{36pt}<{\centering}}
    \hline
    
    \multicolumn{8}{c}{\rule{0pt}{11pt}Text Images in Train Set of SR-RAW}\\
    \hline

    \rule{0pt}{11pt}Focal Length & 240mm & 150mm & 100mm & 70mm & 50mm & 35mm & 24mm\\
    \hline

    \rule{0pt}{11pt}Original Image Number & 393 & 393 &393 & 393 & 393 & 393 & 365    \\

   \rule{0pt}{11pt}Text Box Number & 9160 & 9160  & 9160 & 9160  & 9160 & 9160 & 8119  \\
  
    \hline
    \rule{0pt}{11pt}Recognition Accuracy & {81.4$\%$} & 
    {69.0$\%$} & 
    {52.1$\%$} & 
    {38.6$\%$} &
    {25.7$\%$} &
    {15.0$\%$} &
    {7.9$\%$} \\
    \hline

    \multicolumn{8}{c}{\rule{0pt}{11pt}Text Images in Test Set of SR-RAW}\\
    \hline

    \rule{0pt}{11pt}Original Image Number & 50 & 50 & 50 & 50 & 50 & 50 & 48    \\

   \rule{0pt}{11pt}Text Box Number & 1734 & 1734  & 1734 & 1734 & 1734 & 1734 & 1630  \\
   \hline

    \rule{0pt}{11pt}Recognition Accuracy & {72.4$\%$} & 
    {65.3$\%$} & 
    {54.4$\%$} & 
    {35.6$\%$} &
    {23.2$\%$} &
    {13.6$\%$} &
    {6.3$\%$} \\
    \hline

    \end{tabular}

%% file: tables/realsr.tex
\begin{tabular}{l|p{30pt}|p{30pt}|p{30pt}}
\hline
\multicolumn{4}{c}{\rule{0pt}{14pt}Text Images in RealSR}         \\ \hline
\rule{0pt}{14pt}Focal Length         & 105mm & 50mm & 28mm \\ \hline
\rule{0pt}{14pt}Original Number      &  115
  & 115   &  115 \\ \hline
\rule{0pt}{14pt}Text Box Number      &     6048 &   6048  &   6048  \\ \hline
\rule{0pt}{14pt}Recognition Accuracy &    75.0$\%$
 &  46.1$\%$  & 16.7$\%$
    \\ \hline
\end{tabular}

%% file: tables/divided_acc.tex
    \begin{tabular}{p{90pt}|p{36pt}<{\centering}p{36pt}<{\centering}p{36pt}<{\centering}p{36pt}<{\centering}p{36pt}<{\centering}p{36pt}<{\centering}p{36pt}<{\centering}}
    \hline
    
     \multicolumn{8}{c}{\rule{0pt}{12pt}Recognition Accuracy of images in different height}\\
    \hline

    \rule{0pt}{12pt}Height(pixels) & 128$-$ & 64$-$128 & 32$-$64 & 16$-$32 & 8$-$16 & 4$-$8 & 0$-$4 \\
    \hline

    \rule{0pt}{12pt}Number & 1586 & 3957   &   9663 & 14862 & 15434 & 11866 &5711   \\

    \rule{0pt}{12pt}Recognition Accuracy & {75.2$\%$} & 
    {84.2$\%$} & 
    {84.6$\%$} & 
    {79.5$\%$} &
    {39.1$\%$} &
    {2.8$\%$} &
    {0.3$\%$} \\
    \hline

    \end{tabular}